\documentclass[11pt]{article}

\usepackage[final]{acl}

\usepackage[utf8]{inputenc}      %
\usepackage[T1]{fontenc}         %
\usepackage{times}               %
\usepackage{latexsym}            %
\usepackage{inconsolata}         %

\usepackage{microtype}           %

\usepackage{url}                 %
\usepackage{breakurl}            %
\usepackage[breaklinks]{hyperref}%

\usepackage{xcolor}
\usepackage{tcolorbox}
\tcbuselibrary{breakable}

\usepackage{amsmath}
\usepackage{amssymb}
\usepackage{mathtools}
\usepackage{amsthm}
\usepackage{amsfonts}
\usepackage{braket}
\usepackage{nicefrac}
\usepackage{algorithm}
\usepackage{algorithmic}

\usepackage{booktabs}
\usepackage{siunitx}
\usepackage{multirow}
\usepackage{makecell}

\usepackage{graphicx}
\usepackage{subfigure}
\usepackage{tikz}
\usetikzlibrary{positioning,arrows.meta}
\usepackage{stfloats}
\usepackage{placeins}

\usepackage[capitalize,noabbrev]{cleveref}

\theoremstyle{plain}
\newtheorem{theorem}{Theorem}[section]

\newtheorem{lemma}[theorem]{Lemma}

\theoremstyle{definition}

\newtheorem{assumption}[theorem]{Assumption}
\theoremstyle{remark}
\newtheorem{remark}[theorem]{Remark}

\newcommand{\scheme}[1]{\mbox{\textsc{#1}}}
\newcommand{\llamathreetwo}{Llama-3.2-3B}
\newcommand{\llamathreeone}{Llama-3.1-8B}
\newcommand{\gemmasevenb}{Gemma-7B}
\newcommand{\falconsevenb}{Falcon-7B}
\newcommand{\mpnet}{paraphrase-multilingual-mpnet-base-v2}
\newcommand{\efive}{multilingual-e5-base}
\newcommand{\bertlargeuncased}{BERT-large-uncased}
\newcommand{\gptfouromini}{GPT-4o-mini}
\newcommand{\gptfourominidated}{GPT-4o-mini-2024-07-18}
\newtcolorbox{promptbox}[1]{
    title={#1},
    colback=black!2,
    colframe=black!45,
    colbacktitle=black!7,
    coltitle=black,
    fonttitle=\small\bfseries,
    fontupper=\small\ttfamily,
    boxrule=0.5pt,
    arc=1pt,
    left=6pt,
    right=6pt,
    top=5pt,
    bottom=5pt,
    toptitle=3pt,
    bottomtitle=3pt,
    before skip=0.75\baselineskip,
    after skip=0.75\baselineskip,
    before upper={\setlength{\parindent}{0pt}\raggedright}
}
\newcommand{\promptplaceholder}[1]{\textbf{\{#1\}}}

\title{Robust Text Watermarking for Large Language Models\\via Dual Semantic Embeddings}

\author{
    Jonas Schäfer \qquad Cezary Pilaszewicz \qquad Gerhard Wunder \\
    Department of Mathematics and Computer Science \\ Freie Universität Berlin \\ Berlin, Germany \\
    \small{
        \textbf{Correspondence:} \href{mailto:jonas.schaefer2@fu-berlin.de}{jonas.schaefer2@fu-berlin.de}
    }
}

\begin{document}
\maketitle
\begin{abstract}
This work presents Dual-Embedding Watermarking (\scheme{DEW}), a semantic watermarking scheme for large language models (LLMs) that leverages contextual and token-level embeddings to enhance robustness against paraphrasing and translation.\ \scheme{DEW} utilizes a signal-processing methodology, applying algebraic vector-space operations to token and context embeddings to derive a watermark signal that degrades gracefully under semantic shifts. The method obfuscates the watermark by projecting embedding vectors through pseudo-random matrices seeded with a secret key. Experimental results show that dual-embedding watermarking can offer state-of-the-art robustness, particularly against translation, while incurring relatively low computational overhead compared with other semantic schemes. At lower watermark strength, \scheme{DEW} also maintains competitive text quality, suggesting that dual-embedding signals provide a promising substrate for robust semantic watermarking.

\end{abstract}

\section{Introduction}
\label{sec:introduction}
Large language models (LLMs) have rapidly emerged as powerful tools that generate text with human-like fluency and are widely used in creative writing, programming, and conversational agents. However, as these models advance, distinguishing LLM-generated from human-authored text becomes increasingly challenging, with profound implications for trust, misinformation, and content attribution.

Inference-time watermarking has recently gained significant attention in both research and policy discussions. These methods introduce a hidden statistical signal into the text during generation, which a corresponding algorithm can detect.

Surface-level watermarks dynamically modify the generation process according to a predefined scheme, such as only sampling from a specific partition of the vocabulary~\cite{pmlr-v202-kirchenbauer23a} or adding keyed exponential noise to logits~\cite{aaronson_watermarking_2022}. Distribution-preserving schemes have been designed to leave the output distribution unchanged under stated conditions~\cite{hu_unbiased_2024,wu_resilient_2024,wu_distortion-free_2024}. Recent work has demonstrated that incorporating text semantics into the watermark signal computation improves resiliency to semantically invariant text modifications, typically at the cost of decreased text quality and/or increased computational overhead~\cite{liu_semantic_2024, hou_semstamp_2024}.

Despite recent advances, paraphrasing and translation remain major challenges in detecting LLM-generated text. Furthermore, many watermarking schemes introduce patterns detectable by third parties, making them susceptible to reverse-engineering attacks~\cite{jovanovic_watermark_2024}. These patterns enable adversaries not only to identify the presence of a watermark but also to remove it systematically.

LLM-generated text is commonly rewritten to meet application-specific demands or stylistic preferences. As a result, text watermarking faces persistent challenges, including threats to content authenticity, diminished trust in AI systems, regulatory ambiguities, and difficulties in legal enforcement. This motivates watermarking methods that remain detectable after such modifications while minimizing the misclassification of human-authored texts.

In this work, we investigate dual-embedding watermark signals that incorporate not only contextual semantics but also the semantics of candidate tokens. Unlike prior methods~\cite{hou_semstamp_2024}, our approach achieves this with low computational overhead and allows the signal to degrade smoothly under semantic shifts.

We present \emph{Dual-Embedding Watermarking} (\scheme{DEW}), which combines two semantic embedding models to compute per-token watermark biases based on the cosine similarity between token and context embeddings. This procedure adds zero-centered, pseudo-random noise to the LLM-computed logits. During detection, watermarked tokens exhibit significantly higher signal scores than unwatermarked tokens in expectation. Crucially, since the variation in watermark signals depends on the differences in token embedding vectors, semantically similar tokens receive similar signals. This property supports robustness to translation and has been largely overlooked in prior work, aside from a few exceptions~\cite{he_can_2024, hou_semstamp_2024}.

Our results show that dual-embedding watermarking can offer state-of-the-art robustness, particularly after translation. In our evaluation, \scheme{DEW} achieves the highest post-translation detection rates and remains competitive after paraphrasing. It also incurs relatively low computational overhead compared with other semantic schemes, while its lower-strength configuration maintains competitive text quality. Together, these results position dual-embedding watermarking as a promising substrate for robustness to meaning-preserving transformations.

The remainder of this paper is organized as follows: Section~\ref{sec:related-work} reviews related work, Section~\ref{sec:methodology} details the methodology, Section~\ref{sec:experiments} presents experimental results, Section~\ref{sec:conclusion} concludes with implications, and Section~\ref{sec:limitations} outlines limitations and future directions.

\section{Related Work}
\label{sec:related-work}
Text watermarking is a special case of linguistic steganography that embeds a hidden signal in a passage of text. LLM watermarks are commonly evaluated along three core dimensions: \emph{detectability}, requiring verifiability at low false-positive rates; \emph{secrecy}, requiring no easily detectable artifacts; and \emph{robustness}, requiring evasion to substantially modify the watermarked text, especially its semantics~\cite{kuditipudi_robust_2024}. These goals are inherently in tension: stronger detectability can reduce secrecy or robustness, while stronger secrecy, including distortion-freeness, can make detection harder. Effective LLM watermarks should also be computationally efficient and compatible with standard autoregressive decoding.

Before LLMs, text watermarking largely relied on rule-based transformations such as synonym substitution~\cite{topkara_hiding_2006} and paraphrasing~\cite{atallah_natural_2003}. Because such methods use fixed substitutions, they systematically alter text statistics, making the watermark easier to detect and remove~\cite{tang_science_2024,ziegler_neural_2019}. Recent LLM watermarking research instead focuses on inference-time schemes, which embed the signal directly during generation by modifying the model's token-selection process, but typically require access to model logits. Consequently, they can typically be disabled in locally hosted models.

\subsection{Surface-level Watermarks}
Most LLM watermarks operate at the surface level, injecting the signal based on token identities or exact token contexts without explicitly modeling semantics. These methods are simple and inexpensive, but exact context dependence makes their signals vulnerable to local edits. \citet{pmlr-v202-kirchenbauer23a,kirchenbauer_reliability_2024} propose a scheme, referred to here as \scheme{KGW}, that hashes the preceding $k$ tokens to pseudo-randomly partition the vocabulary into green and red lists and then boosts green-list logits. Parallel unpublished work by \citet{aaronson_watermarking_2022}, referred to as \scheme{EXP}, similarly hashes the previous $k$ tokens but samples using keyed exponential noise and Perturb-and-MAP decoding~\cite{papandreou_perturb-and-map_2011}. Both schemes bias the distribution toward subsets of $k$-grams~\cite{kuditipudi_robust_2024,wu_distortion-free_2024}, yielding a trade-off: larger $k$ improves secrecy by reducing repeated contexts, whereas smaller $k$ improves robustness by making local edits less disruptive. In the limiting case $k=0$, \scheme{KGW} becomes \scheme{Unigram}~\cite{zhao_provable_2024}, which is highly robust but vulnerable to reverse-engineering attacks~\cite{jovanovic_watermark_2024}. \citet{dathathri_scalable_2024} instead propose \scheme{SynthID}, which uses Tournament Sampling to optimize a secret statistical watermark score and also provides a distortion-free mode with reduced detectability. \scheme{MorphMark}~\cite{wang-etal-2025-morphmark} adapts green-list watermark strength to the model's cumulative probability mass on the green list.

Distortion-free and distribution-preserving watermarks aim to improve secrecy by avoiding changes to the output distribution. In this sense, \scheme{EXP} is distortion-free only when $k$ is large enough to avoid repeated contexts. \scheme{UnbiasedWM}~\cite{hu_unbiased_2024} uses inverse-transform sampling and permutation-based reweighting to integrate a watermark without altering token probabilities, but its detection requires token logits and ideally an approximate reconstruction of the prompt~\cite{wu_resilient_2024}. \scheme{DiPmark}~\cite{wu_resilient_2024} instead uses a distribution-preserving reweighting function to increase the probability mass assigned to a pseudo-random token set. Both \scheme{UnbiasedWM} and \scheme{DiPmark} are provably distortion-free in the absence of watermark key collisions~\cite{wu_distortion-free_2024}.

\subsection{Semantic Watermarks}
Semantic watermarks are motivated by the limited robustness of surface-level schemes against meaning-preserving transformations such as paraphrasing and translation. Rather than relying only on token hashes, they condition the watermark signal, its parameters, or its training objective on semantic representations. Some semantic watermarks, including \scheme{DEW}, also make each candidate token's signal depend on that token's semantics. This distinction is important because context-level semantics can stabilize the signal under paraphrasing, whereas candidate-token semantics make synonym substitutions and translations more likely to preserve token-level evidence.

\scheme{TS}~\cite{huo_token-specific_2024} extends the green-list paradigm by learning token-specific vocabulary split ratios and green-list logit biases from the preceding-token embedding. Detection remains \scheme{KGW}-like via a one-sided $z$-test adjusted for varying split ratios. While \scheme{TS} improves detectability and semantic coherence over fixed-parameter green-list schemes, it does not directly assign similar watermark signals to semantically related candidate tokens.

\scheme{ATW}~\cite{liu_adaptive_2024} combines entropy-gated insertion with semantic logit scaling. It leaves low-entropy decoding steps unmodified and, at selected high-entropy steps, maps embeddings of the preceding text to a logits-scaling vector. Detection approximates a likelihood-ratio test over the tokens selected by the same entropy criterion. Compared with \scheme{DEW}, \scheme{ATW} conditions the signal on context semantics but does not explicitly couple candidate-token scores to candidate-token semantics.

\scheme{SIR}~\cite{liu_semantic_2024} uses an auxiliary LLM to embed the preceding context and transforms these embeddings into watermark logits with a neural network trained to preserve semantic similarity while maintaining diversity and unbiasedness. This makes the watermark more stable under semantically invariant edits, but the signal is primarily context-conditioned and requires an auxiliary learned mapping in addition to the host LLM.

\scheme{X-SIR}~\cite{he_can_2024} extends \scheme{SIR} by clustering semantically similar tokens and assigning a shared watermark bias within each cluster, making it the closest prior work to \scheme{DEW} because it incorporates candidate-token semantics during signal computation.

\scheme{SemStamp}~\cite{hou_semstamp_2024} operates at sentence granularity: it embeds each generated sentence and uses rejection sampling to output only sentences whose embedding falls into an allowed locality-sensitive hashing (LSH) partition. Its successor, \scheme{$k$-SemStamp}~\cite{hou_k-semstamp_2024}, replaces LSH with $k$-means clustering, improving the robustness and sampling efficiency reported by its authors. We therefore evaluate \scheme{$k$-SemStamp} rather than \scheme{SemStamp}. However, \scheme{$k$-SemStamp} requires domain-specific initialization and remains considerably more expensive than token-level insertion: in our setup, generation takes approximately twelve times as long as with \scheme{DEW} (Table~\ref{tab:comp-efficiency}). Moreover, in our testing, 2.1\% of sentence positions reached the 100-trial rejection-sampling limit and used the final candidate without enforcing the acceptance condition.

Additional recent watermarking methods not included in our evaluation are discussed in Appendix~\ref{app:additional-related-work}.

\section{Methodology}
\label{sec:methodology}
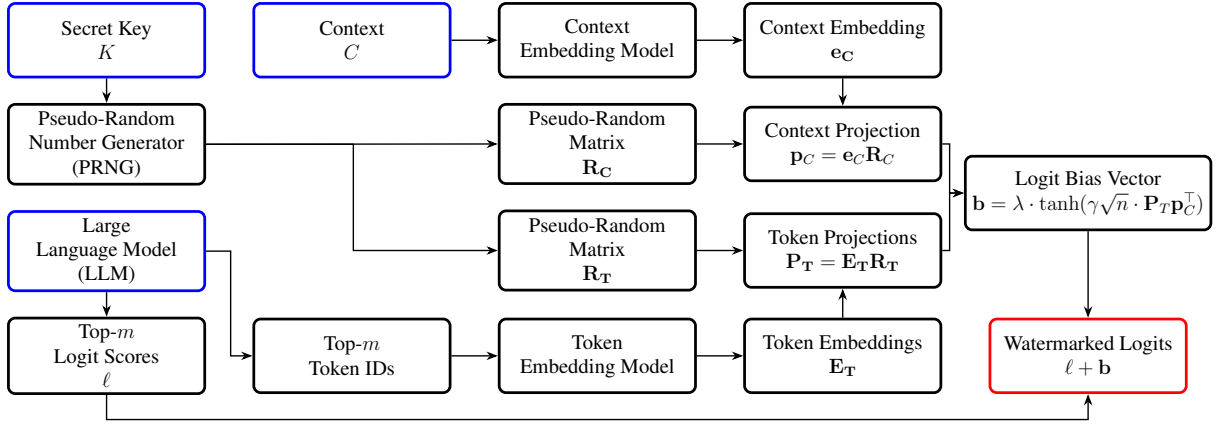
\begin{figure*}[t]
    \centering
    \resizebox{\textwidth}{!}{\begin{tikzpicture}[
    node distance={5cm}, 
    every node/.style={
        draw, 
        rounded corners, 
        align=center, 
        minimum width=4cm, 
        minimum height=1.5cm,
    }
]
\large
\node[draw=blue, ultra thick] (secret_key) {Secret Key\\$K$};
\node[draw=blue, ultra thick, right of=secret_key] (context) {Context\\$C$};
\node[ultra thick, right of=context] (context_embedding_model) {Context\\Embedding Model};
\node[ultra thick, right of=context_embedding_model] (context_embedding) {Context Embedding\\$\mathbf{e_C}$};

\node[ultra thick, below=0.5cm of secret_key] (prng) {Pseudo-Random\\Number Generator\\(PRNG)};
\node[ultra thick, below=0.5cm of context_embedding_model] (random_matrix_rc) {Pseudo-Random\\Matrix\\$\mathbf{R_C}$};
\node[draw, ultra thick, right of=random_matrix_rc] (context_projection) {Context Projection\\$\mathbf{p}_C = \mathbf{e}_C \mathbf{R}_C$};

\node[draw=blue, ultra thick, below=0.5cm of prng] (llm) {Large\\Language Model\\(LLM)};
\node[ultra thick, below=0.5cm of random_matrix_rc] (random_matrix_rt) {Pseudo-Random\\Matrix\\$\mathbf{R_T}$};
\node[draw, ultra thick, right of=random_matrix_rt] (token_projections) {Token Projections\\$\mathbf{P_T} = \mathbf{E_T} \mathbf{R_T}$};

\node[ultra thick, below=0.5cm of llm] (top_m_scores) {Top-$m$\\Logit Scores\\$\mathbf{\ell}$};
\node[ultra thick, right of=top_m_scores] (top_m_ids) {Top-$m$\\Token IDs};
\node[ultra thick, right of=top_m_ids] (token_embedding_model) {Token\\Embedding Model};
\node[ultra thick, right of=token_embedding_model] (token_embeddings) {Token Embeddings\\$\mathbf{E_T}$};

\node[ultra thick, right of=context_projection, yshift=-1cm] (logit_bias_vector) {Logit Bias Vector\\$\mathbf{b} = \lambda \cdot \tanh(\gamma\sqrt{n}\cdot\mathbf{P}_T \mathbf{p}_C^\top)$};

\node[draw=red, ultra thick, right of=token_embeddings] (watermarked_logits) {Watermarked Logits\\$\mathbf{\ell} + \mathbf{b}$};

\draw[-{Stealth}, thick] (secret_key) -- (prng);
\draw[-{Stealth}, thick] (prng.east) -- (random_matrix_rc.west);
\draw[-{Stealth}, thick] (prng.east) -- +(3,0) |- (random_matrix_rt.west);
\draw[-{Stealth}, thick] (context) -- (context_embedding_model);
\draw[-{Stealth}, thick] (context_embedding_model) -- (context_embedding);
\draw[-{Stealth}, thick] (context_embedding) -- (context_projection);
\draw[-{Stealth}, thick] (random_matrix_rc) -- (context_projection);
\draw[-{Stealth}, thick] (random_matrix_rt) -- (token_projections);
\draw[-{Stealth}, thick] (llm) -- (top_m_scores);
\draw[-{Stealth}, thick] (llm.east) -- +(0.5,0) |- (top_m_ids.west);
\draw[-{Stealth}, thick] (top_m_ids) -- (token_embedding_model);
\draw[-{Stealth}, thick] (token_embedding_model) -- (token_embeddings);
\draw[-{Stealth}, thick] (token_embeddings) -- (token_projections);
\draw[-{Stealth}, thick] (context_projection.east) -- +(0.15,0) |- (logit_bias_vector.west);
\draw[-{Stealth}, thick] (token_projections.east) -- +(0.15,0) |- (logit_bias_vector.west);
\draw[-{Stealth}, thick] (logit_bias_vector) -- (watermarked_logits);
\draw[-{Stealth}, thick] (top_m_scores.south) -- +(0,-0.5) -| (watermarked_logits.south);

\end{tikzpicture}}
    \caption{An illustration of the \scheme{DEW} insertion procedure for a single generation step. Previously generated tokens ($C$) are jointly embedded, while the top-$m$ candidate token embeddings are computed separately. All embeddings are projected for obfuscation, and the dot product of the projections is added to the original logits as token-specific watermark biases. We sample from the updated logits. Inputs are highlighted in {\color{blue}blue}, and the output watermarked logits in {\color{red}red}. For conciseness, we omitted normalization, whitening, orthonormalization, and standardization from the diagram.}
    \label{fig:watermarking_scheme}
\end{figure*}

LLMs have a vocabulary $\mathcal{V}$ containing words or word fragments (\emph{tokens}). Given an input sequence $\mathbf{x} = (x_1, \ldots, x_{t-1})$, the model computes a probability distribution over $\mathcal{V}$ by producing a set of \emph{logits} $\ell$, where each logit represents the unnormalized log-probability of the corresponding token. Each token $x_t$ is selected by sampling from this distribution or using a decoding method such as beam search. This process repeats until the LLM generates an end-of-sequence token or reaches a maximum text length.

Inference-time watermarking schemes modify probability distributions by either manipulating the sampling process or by directly adjusting the distribution, as in this work, where watermark biases are added to the candidate token logits during text generation (\emph{watermark insertion}, Section~\ref{subsec:insertion}). To introduce secrecy, this process generally employs a pseudo-random number generator (PRNG) that modifies the signal using a secret key known only to the model provider. Most schemes also require this key to determine whether a candidate text contains the watermark, a procedure known as \emph{watermark detection} (Section~\ref{subsec:detection}).

To improve text diversity and, in turn, secrecy, the embedded signal is typically made dependent on a sliding window of directly preceding tokens (the \emph{watermark context}) by hashing them along with the secret key. However, due to the cryptographic nature of the hash function, even minor changes in the context yield statistically independent signals. For this reason, the robustness of such schemes decreases with larger watermark context widths, although text diversity and watermark secrecy improve.

Semantic watermarks enhance robustness against semantically invariant modifications, such as paraphrasing and translation. These schemes use a numeric representation of the context semantics to assign the same vocabulary partitioning to semantically similar contexts. This representation is commonly obtained through \emph{embedding models}, which compute vector representations of token sequences. These models are trained, for example, via contrastive learning, to map semantically similar texts to nearby points in the embedding space. Semantic watermarking leverages this property by making the watermark signal contingent on the embedding vector.

Although semantic watermarks offer improved robustness to semantically invariant changes in the watermark context, most schemes do not consider inter-token semantic similarity when calculating the bias distributions. For this reason, substituting a token with a synonym has a high chance of removing the signal embedded in that token. \scheme{DEW} computes separate semantic embeddings for the context and for each candidate token to assign similar biases to tokens with close embedding vectors. Additionally, the signal carried by each token smoothly degrades with semantic shifts in either the context or the token itself, further improving robustness.

\subsection{Watermark Insertion}
\label{subsec:insertion}

\subsubsection{Setup}
In addition to the LLM, \scheme{DEW} incorporates two embedding models. The token embedding model $M_T$ maps individual tokens to $d_T$-dimensional vectors. Similarly, the context embedding model $M_C$ maps token sequences of arbitrary length to $d_C$-dimensional vectors. In our experiments, $M_T$ is the host LLM's input-embedding table and $M_C$ is a multilingual sentence encoder; robustness depends on the suitability of both embedding spaces (Appendix~\ref{app:ablation-study}).

To initialize the algorithm, a secret key $K$ is employed to seed a cryptographically secure PRNG. For the sake of simplicity and efficiency in our experiments, we opted for the default non-secure PyTorch Philox PRNG. This generator randomly samples from the standard normal distribution to produce two matrices $\mathbf{R}_T \in \mathbb{R}^{d_T \times n}$ and $\mathbf{R}_C \in \mathbb{R}^{d_C \times n}$. Through random projections, these matrices obfuscate the embedding vectors while preserving distances.

We further introduce the value $n$, which we call the \emph{projection dimensionality}. While the embedding models determine $d_T$ and $d_C$, $n$ is a tunable hyperparameter controlling the dimension of the random-projection space. We conservatively set $n=\max(d_T,d_C)$ in this work. However, by the Johnson-Lindenstrauss lemma~\cite{10.1090/conm/026/737400}, one can often choose a significantly smaller $n$ while approximately preserving distances. Furthermore, in Appendix~\ref{app:orthogonal_construction}, we propose an optional block-wise orthogonal construction of $R_T$ and $R_C$ that is \emph{guaranteed} to preserve angles between embedding vectors while still obfuscating them through pseudo-random rotations.

\subsubsection{Semantic Extraction}
At each generation step, the LLM computes the logits $\mathbf{\ell} \in \mathbb{R}^{|\mathcal{V}|}$ as usual. We use $M_T$ to embed all subsequent candidate tokens. In practice, to reduce computational overhead, it is often sufficient to consider only the top $m\in\mathbb{N}$ tokens with the highest scores in $\mathbf{\ell}$, yielding an embedding matrix $\mathbf{E}_T \in \mathbb{R}^{m \times d_T}$. Another option is to apply nucleus sampling, which dynamically selects the smallest set of tokens whose cumulative probability exceeds a specified threshold. Each row of $\mathbf{E}_T$ is an embedding vector in $\mathbb{R}^{d_T}$ associated with one of the $m$ highest-scoring candidate tokens.

Optionally, a whitening transformation can be applied to the token embeddings to ensure isotropy (uniformity in all directions) in the embedding space. For various applications, whitening generally makes embedding similarity metrics more meaningful and consistent across dimensions~\cite{huang_whiteningbert_2021, diera_isotropy_2025}. Since sequence embeddings are typically derived by pooling individual token embeddings, it is also feasible to apply whitening before pooling. However, we only whiten token embeddings in this study.

\begin{algorithm}[t]
\caption{\scheme{DEW} Watermark Insertion (Single Step)}
\label{alg:watermark_insertion}
\begin{algorithmic}[1]

\REQUIRE LLM logits $\mathbf{\ell} \in \mathbb{R}^{|\mathcal{V}|}$, 
  watermark context $\mathbf{c} = (x_{t-k}, \ldots, x_{t-1})$, 
  token embedding model $M_T$, 
  context embedding model $M_C$, 
  secret key $K$, 
  top-$m$ candidate count,
  watermark strength $\lambda$,
  saturation factor $\gamma$,
  projection dimensionality $n$.

\ENSURE Watermarked logits $\mathbf{\ell'}$

\STATE Use $K$ to seed a PRNG \hfill\textit{(only once per session; can be cached)}
\STATE Regenerate (or recall) $\mathbf{R}_T \in \mathbb{R}^{d_T \times n}$ and $\mathbf{R}_C \in \mathbb{R}^{d_C \times n}$

\STATE \textbf{Compute projected context embedding:}
\STATE \quad 
  $\mathbf{e}_C \gets M_C(\mathbf{c}) \in \mathbb{R}^{d_C}$
\STATE \quad Normalize $\mathbf{e}_C$
\STATE \quad 
  $\mathbf{p}_C \gets \operatorname{normalize}(\mathbf{e}_C\,\mathbf{R}_C) \in \mathbb{R}^n$

\STATE \textbf{Compute (or recall) projected token embeddings:}
\STATE \quad Let $\mathcal{T} \subseteq \mathcal{V}$ be the set of top-$m$ tokens from $\mathbf{\ell}$
\STATE \quad $\mathbf{E}_T \gets M_T(\mathcal{T}) \in \mathbb{R}^{m \times d_T}$
\STATE \quad \textit{Optional}: Apply whitening to $\mathbf{E}_T$
\STATE \quad Normalize rows of $\mathbf{E}_T$
\STATE \quad $\mathbf{P}_T \gets \operatorname{row\_normalize}(\mathbf{E}_T\,\mathbf{R}_T) \in \mathbb{R}^{m \times n}$

\STATE \textbf{Compute biases and add them to logits:}
\STATE \quad
  $\mathbf{b} \gets \lambda \cdot \tanh\!\Bigl(\gamma\sqrt{n}\cdot\mathbf{P}_T\mathbf{p}_C^\top\Bigr) \in \mathbb{R}^m$
\STATE \quad
  Set $\mathbf{\ell'}$ by adding each entry of $\mathbf{b}$ to its corresponding position in $\mathbf{\ell}$

\STATE \algorithmicoutput \ $\mathbf{\ell'}$ \hfill\textit{(watermarked logits for the next token)}

\end{algorithmic}
\end{algorithm}

\subsubsection{Obfuscation}
Next, we normalize the rows of $\mathbf{E}_T$ and multiply the result by $\mathbf{R}_T$, applying a random linear transformation to each embedding vector for obfuscation. Since the token and context embedding models are fixed (and potentially public), obfuscating the embeddings is essential to enable secrecy. We achieve this through the secret linear transformations $\mathbf{R}_T$ and $\mathbf{R}_C$. The same process is applied to the watermark context (for example, the $k$ preceding tokens), yielding an embedding vector $\mathbf{e}_C \in \mathbb{R}^{d_C}$ and projection vector $\mathbf{p}_C = \mathbf{e}_C \mathbf{R}_C \in \mathbb{R}^n$. Notably, all token embeddings and their projections can be precomputed offline.

\subsubsection{Bias Computation}
We then calculate the logit bias vector $\mathbf{b}$ by taking the dot product of the context projection vector with each projected token embedding vector. Since both vectors are normalized, their dot product equals the cosine similarity, which ranges from $-1$ to $1$ and quantifies the cosine of the angle between them. This value reflects the degree of alignment, with $1$ indicating perfect alignment, $-1$ perfect opposition, and $0$ orthogonality. Since $\mathbf{E}_T$ and $\mathbf{e}_C$ may originate from different models and are independently obfuscated through projection, the cosine similarity lacks a direct interpretive meaning. Nevertheless, it provides a useful keyed semantic alignment signal: small changes in token or context embeddings induce controlled changes in the projected cosine score, so semantically similar continuations tend to receive similar biases. Under an isotropic spherical null model, this score is symmetric around $0$ with a known Beta-type distribution (Appendix~\ref{app:math}), so text generated independently of $K$ attains an expected score of zero. We derive the exact null distribution of the alignment score $\mathbf{p}_T \mathbf{p}_C^\top$ (Beta-type) and its high-dimensional Gaussian approximation, and use it to motivate the $\sqrt{n}$ scaling and false-positive calibration.%

The variance of the dot product of two random vectors uniformly distributed on the unit sphere depends on the dimension of said vectors. In our idealized null model, the dot product of a spherical vector with any fixed unit vector has variance $1/n$. Therefore, multiplying the dot products by $\sqrt{n}$ yields an approximately unit-variance baseline score. In practice, the spherical model is an analytic baseline rather than an exact description of natural-language token statistics. We therefore use it to motivate the $\sqrt{n}$ scaling and complement it with empirical thresholding when reporting fixed-FPR detection results.

Finally, the dot products are passed through the $\tanh$ function to compute the bias vector $\mathbf{b}$:
\begin{equation}
\label{eq:bias-insertion}
    \mathbf{b} = \lambda \cdot \tanh(\gamma\sqrt{n}\cdot\mathbf{P}_T \mathbf{p}_C^\top) \in \mathbb{R}^m
\end{equation}
Here, $\lambda$ is a hyperparameter to control the watermark signal strength. Moreover, $\tanh(\cdot)$ denotes the element-wise application of the hyperbolic tangent function. Using a non-linear activation function such as $\tanh$ facilitates smooth clipping of the pre-scaling biases, mitigating the impact of outliers on text quality. By scaling the argument of $\tanh$ by a hyperparameter $\gamma\in\mathbb{R}^+_*$, the clipping level can be adjusted: larger scaling factors accentuate saturation, while smaller factors preserve a broader dynamic range of biases. Higher saturation results in more tokens receiving extreme bias values, approaching $-\lambda$ or $\lambda$. This behavior is reminiscent of the green/red list \scheme{KGW} watermark~\cite{pmlr-v202-kirchenbauer23a}, though \scheme{KGW} does not involve assigning negative biases to logits. Finally, we insert $\mathbf{b}$ into the corresponding $m$ positions of $\mathbf{\ell}$ and sample the next token from the updated logits. We describe the watermark insertion procedure for one generation step in Algorithm~\ref{alg:watermark_insertion} and illustrate it in Figure~\ref{fig:watermarking_scheme}.

\subsection{Watermark Detection} \label{subsec:detection}
The detection procedure mirrors the insertion process. It iterates over a given candidate text document token by token and sums the biases embedded in each token to obtain a document-level watermark score. This score can be thresholded for binary classification, with the threshold tunable to control the FPR.

Specifically, we tokenize the candidate text with the host tokenizer and look up the compatible precomputed projection $\mathbf{p}_T$ for each observed token $x_t$. For the context $\mathbf{c}=(x_{t-k},\ldots,x_{t-1})$, we compute its embedding vector $\mathbf{e}_C=M_C(\mathbf{c})$, normalize it, and project it via $\mathbf{R}_C$ to obtain $\mathbf{p}_C$. Finally, we obtain the token bias score by computing $\mathbf{b}=\lambda\cdot\tanh (\gamma\sqrt{n}\cdot\mathbf{p}_T\mathbf{p}_C^\top)\in\mathbb{R}$, which only differs from Equation~\ref{eq:bias-insertion} in $\mathbf{p}_T$, as we now only compute the score of the observed token, instead of $m$ candidate tokens. Thus, detection requires the host tokenizer, a compatible precomputed projected token table, the context encoder, and the secret key, but no host-model forward pass, logits, hidden states, or live host weights. The full detection procedure is described in Algorithm~\ref{alg:watermark_detection}.

\begin{algorithm}[t]
\caption{\scheme{DEW} Watermark Detection (Single Step)}
\label{alg:watermark_detection}
\begin{algorithmic}[1]

\REQUIRE Observed token $x_t$,
  watermark context $\mathbf{c} = (x_{t-k}, \ldots, x_{t-1})$,
  host tokenizer,
  compatible precomputed projected token table $\mathbf{P}_T$,
  context embedding model $M_C$, 
  secret key $K$,
  watermark strength $\lambda$,
  saturation factor $\gamma$,
  projection dimensionality $n$.

\ENSURE Token-level watermark score $s$

\STATE Use $K$ to seed a PRNG \hfill\textit{(only once per session; can be cached)}
\STATE Regenerate (or recall) $\mathbf{R}_C \in \mathbb{R}^{d_C \times n}$

\STATE \textbf{Compute projected context embedding:}
\STATE \quad 
  $\mathbf{e}_C \gets M_C(\mathbf{c}) \in \mathbb{R}^{d_C}$
\STATE \quad Normalize $\mathbf{e}_C$
\STATE \quad 
  $\mathbf{p}_C \gets\operatorname{normalize}(\mathbf{e}_C\,\mathbf{R}_C) \in \mathbb{R}^n$

\STATE \textbf{Look up projected token embedding:}
\STATE \quad $\mathbf{p}_T \gets \mathbf{P}_T[x_t] \in \mathbb{R}^n$

\STATE \textbf{Compute watermark score:}
\STATE \quad
  $s \gets \lambda \cdot \tanh\!\Bigl(\gamma\sqrt{n}\cdot\mathbf{p}_T\mathbf{p}_C^\top\Bigr) \in \mathbb{R}$

\STATE \algorithmicoutput \ $s$ \hfill\textit{(token-level watermark score)}
 
\end{algorithmic}
\end{algorithm}

The detection procedure can be formalized as a statistical hypothesis test (Appendix~\ref{app:math:test}) to control FPRs rigorously and improve interpretability. The resulting empirical score distributions match the Gaussian baseline derived in Appendix~\ref{app:math:dist}. For comparability, we report TPRs at fixed FPRs in Section~\ref{sec:experiments} using the standard empirical thresholding procedure from MarkLLM~\cite{pan_markllm_2024}.

\section{Experiments}
\label{sec:experiments}
\subsection{Language Models and Hyperparameters}
\label{subsec:lms-hyperparams}
We use \llamathreetwo{}~\cite{grattafiori_llama_2024} for all main experiments and additionally evaluate \gemmasevenb{}~\cite{DBLP:journals/corr/abs-2403-08295} and \falconsevenb{}~\cite{DBLP:journals/corr/abs-2311-16867} in Appendix~\ref{app:additional-llms}. We generate text via multinomial sampling. To enhance text diversity, we apply a four-gram blocking constraint. This ensures that no four-token sequence that has already been generated can be repeated.

As \scheme{DEW}'s default semantic context embedding model $M_C$, we choose \mpnet{}~\cite{reimers_sentence-bert_2019} ($d_C=768$) due to its multilingual paraphrase robustness. We obtain the token embeddings from the word embedding layer of the underlying LLM (for \llamathreetwo{}, $d_T=3\,072$).

As hyperparameters for \scheme{DEW}, we use $m=32$, $n=3\,072$, and $\gamma=0.5$ throughout all experiments with \llamathreetwo{}. Further, we apply whitening to token embeddings and take the $\tanh$ of the bias scores before scaling by $\lambda$. As we observe no significant improvement from applying orthonormalization to the random matrices with our specific models, we omit this step from the evaluation.

In Table~\ref{tab:results}, \scheme{DEW} uses \mpnet{} with context width $k=3$. We evaluate watermark strengths $\lambda=2.0$ and $\lambda=2.5$; increasing $\lambda$ strengthens robustness but harms text quality. The hyperparameters for all baseline watermarking schemes are provided in Appendix~\ref{app:configs}.

\subsection{Dataset and Prompt}
For text generation, we use the C4 dataset~\cite{raffel_exploring_2020}, as it is widely employed for evaluating watermarking effectiveness in high-entropy, free-form text generation tasks. From each document, we take the first 30 tokens as the prompt and generate 200 additional tokens as a completion. As counterexamples, we use unwatermarked completions generated by the same model and sampling configuration as the watermarked texts.

\subsection{Detectability and Robustness Analysis}

We report TPR at nominal FPRs of 1\% and 5\% using leave-one-seed-out calibration across three seeds. For each held-out seed, thresholds are calibrated on same-model unwatermarked completions from the other seeds and applied to 500 watermarked and 500 unwatermarked completions. We report the mean and half-range across seeds.

To evaluate robustness against paraphrasing and translation, we prompt \gptfourominidated{} to rewrite the watermarked text while preserving its meaning and tone. Only watermarked positives are transformed, and the English threshold for the corresponding held-out seed is applied to each attacked set of 500 texts. Notably, \gptfouromini{} is substantially more capable than \llamathreetwo{}, which we use for text generation. The exact prompts are presented in Appendix~\ref{app:prompts}.

The nominal FPR is therefore controlled only for English unwatermarked completions from the same host model. Paraphrasing and translation may shift the null-score distribution, so the attacked TPRs in Table~\ref{tab:results} should not be interpreted as operating at 1\% or 5\% FPR in the paraphrased, German, or French domains.

\begingroup
\newcommand{\uncertainty}[1]{{\scriptsize\,\(\pm\)\,#1}}
\newcommand{\best}{\bfseries}
\newcommand{\underlinenum}[3]{{\phantom{\tablenum[table-format=#1,#2]{#3}}\llap{\underline{\num[#2]{#3}}}}}
\newcommand{\categorytpr}[1]{\underlinenum{1.3}{print-zero-integer=false}{#1}}
\newcommand{\categoryppl}[1]{\underlinenum{2.2}{}{#1}}
\newcommand{\categorynps}[1]{\underlinenum{+1.3}{print-zero-integer=false,retain-explicit-plus=true}{#1}}
\begin{table*}[t]
\centering
\caption{True positive rates in unattacked, post-paraphrasing, and post-translation scenarios at nominal English same-model false positive rates of 1 and 5 percent. Values are three-seed means, with half-ranges shown in smaller type. The text quality measures are computed on unmodified watermarked text. PPL represents median perplexity, while NPS measures preference relative to unwatermarked generations. The best scores across all watermarking schemes are highlighted in bold, while the top scores within the other category (semantic/surface-level) are underlined.}
\label{tab:results}
\footnotesize
\setlength{\tabcolsep}{1.4pt}
\renewcommand{\arraystretch}{1.10}
\sisetup{detect-shape=true,detect-weight=true,mode=text,table-number-alignment=center}
\begin{tabular*}{\textwidth}{@{\extracolsep{\fill}}l*{7}{S[table-format=1.3,print-zero-integer=false]@{}l}S[table-format=2.2]@{}lS[table-format=+1.3,print-zero-integer=false,retain-explicit-plus=true]@{}l@{}}
\toprule
& \multicolumn{2}{c}{Unattacked}
& \multicolumn{4}{c}{Paraphrased}
& \multicolumn{4}{c}{German translation}
& \multicolumn{4}{c}{French translation}
& \multicolumn{4}{c}{Text quality} \\
\cmidrule(lr){2-3}\cmidrule(lr){4-7}\cmidrule(lr){8-11}\cmidrule(lr){12-15}\cmidrule(l){16-19}
Watermark
& \multicolumn{2}{c}{TPR@1\%}
& \multicolumn{2}{c}{TPR@1\%} & \multicolumn{2}{c}{TPR@5\%}
& \multicolumn{2}{c}{TPR@1\%} & \multicolumn{2}{c}{TPR@5\%}
& \multicolumn{2}{c}{TPR@1\%} & \multicolumn{2}{c}{TPR@5\%}
& \multicolumn{2}{c}{PPL $\downarrow$} & \multicolumn{2}{c}{NPS $\uparrow$} \\
\midrule
\multicolumn{19}{@{}l}{\textit{Semantic watermarks}} \\
\scheme{DEW} ($\lambda=2.0$)
& 0.986 & \uncertainty{.008} & 0.352 & \uncertainty{.065} & 0.711 & \uncertainty{.019}
& 0.438 & \uncertainty{.066} & 0.857 & \uncertainty{.031} & 0.243 & \uncertainty{.069}
& 0.691 & \uncertainty{.039} & 8.94 & \uncertainty{0} & -0.046 & \uncertainty{.029} \\
\scheme{DEW} ($\lambda=2.5$)
& 0.999 & \uncertainty{.001} & 0.543 & \uncertainty{.051} & 0.831 & \uncertainty{.022}
& \best 0.521 & \uncertainty{.077} & \best 0.904 & \uncertainty{.003} & \best 0.299 & \uncertainty{.070}
& \best 0.754 & \uncertainty{.022} & 9.85 & \uncertainty{.06} & -0.108 & \uncertainty{.029} \\
\scheme{k-SemStamp}
& 0.819 & \uncertainty{.028} & 0.241 & \uncertainty{.024} & 0.467 & \uncertainty{.026}
& 0.001 & \uncertainty{.002} & 0.019 & \uncertainty{.004} & 0.009 & \uncertainty{.001}
& 0.031 & \uncertainty{.004} & {\categoryppl{8.00}} & \uncertainty{0} & \best +0.088 & \uncertainty{.037} \\
\scheme{SIR}
& 0.974 & \uncertainty{.002} & 0.631 & \uncertainty{.033} & 0.838 & \uncertainty{.018}
& 0.211 & \uncertainty{.023} & 0.505 & \uncertainty{.021} & 0.180 & \uncertainty{.023}
& 0.469 & \uncertainty{.028} & 9.79 & \uncertainty{.16} & -0.261 & \uncertainty{.029} \\
\scheme{X-SIR}
& 0.919 & \uncertainty{.014} & 0.586 & \uncertainty{.031} & 0.779 & \uncertainty{.009}
& 0.293 & \uncertainty{.037} & 0.568 & \uncertainty{.016} & 0.213 & \uncertainty{.029}
& 0.484 & \uncertainty{.017} & 9.54 & \uncertainty{.06} & -0.202 & \uncertainty{.040} \\
\scheme{ATW}
& \best 1.000 & \uncertainty{0} & \best 0.646 & \uncertainty{.014} & \best 0.890 & \uncertainty{.009}
& 0.013 & \uncertainty{.002} & 0.091 & \uncertainty{.005} & 0.003 & \uncertainty{.002}
& 0.021 & \uncertainty{.004} & 11.10 & \uncertainty{.13} & -0.048 & \uncertainty{.017} \\
\scheme{TS}
& \best 1.000 & \uncertainty{0} & 0.592 & \uncertainty{.036} & 0.806 & \uncertainty{.019}
& 0.129 & \uncertainty{.016} & 0.306 & \uncertainty{.034} & 0.066 & \uncertainty{.012}
& 0.180 & \uncertainty{.018} & 10.48 & \uncertainty{.06} & -0.125 & \uncertainty{.028} \\
\addlinespace[.6ex]
\multicolumn{19}{@{}l}{\textit{Surface-level watermarks}} \\
\scheme{MorphMark}
& 0.998 & \uncertainty{.002} & 0.415 & \uncertainty{.050} & {\categorytpr{0.673}} & \uncertainty{.018}
& 0.040 & \uncertainty{.019} & 0.119 & \uncertainty{.011} & 0.062 & \uncertainty{.022}
& 0.171 & \uncertainty{.019} & 8.63 & \uncertainty{0} & -0.062 & \uncertainty{.026} \\
\scheme{KGW}
& 0.999 & \uncertainty{.001} & 0.199 & \uncertainty{.018} & 0.394 & \uncertainty{.023}
& 0.027 & \uncertainty{.006} & 0.080 & \uncertainty{.027} & 0.019 & \uncertainty{.007}
& 0.071 & \uncertainty{.014} & 10.46 & \uncertainty{.16} & -0.113 & \uncertainty{.032} \\
\scheme{SynthID}
& 0.999 & \uncertainty{.001} & {\categorytpr{0.423}} & \uncertainty{.043} & 0.607 & \uncertainty{.019}
& 0.023 & \uncertainty{.005} & 0.061 & \uncertainty{.003} & 0.025 & \uncertainty{.005}
& 0.069 & \uncertainty{.002} & \best 6.68 & \uncertainty{.09} & {\categorynps{-0.002}} & \uncertainty{.060} \\
\scheme{DiPmark}
& 0.997 & \uncertainty{.002} & 0.100 & \uncertainty{.003} & 0.267 & \uncertainty{.009}
& 0.008 & \uncertainty{.006} & 0.049 & \uncertainty{.017} & 0.011 & \uncertainty{.002}
& 0.061 & \uncertainty{.002} & 9.15 & \uncertainty{.06} & -0.047 & \uncertainty{.018} \\
\scheme{UnbiasedWM}
& \best 1.000 & \uncertainty{0} & 0.207 & \uncertainty{.037} & 0.353 & \uncertainty{.018}
& {\categorytpr{0.170}} & \uncertainty{.064} & {\categorytpr{0.368}} & \uncertainty{.043} & {\categorytpr{0.073}} & \uncertainty{.028}
& {\categorytpr{0.223}} & \uncertainty{.021} & 9.19 & \uncertainty{.13} & -0.089 & \uncertainty{.039} \\
\bottomrule
\end{tabular*}
\end{table*}
\endgroup

\subsection{Text Quality Analysis}
We report the median \textit{perplexity} (PPL)~\cite{jelinek_perplexity_1977} of the watermarked completions under the more powerful \llamathreeone{} model~\cite{grattafiori_llama_2024}. Perplexity is the exponentiated average negative log-likelihood of the observed token sequence. While it is widely used as a simple proxy for text quality, it can assign favorable scores to repetitive or overconfident outputs that lack diversity or factual accuracy.

To complement PPL, we calculate the \emph{Net Preference Score} (NPS) using \gptfourominidated{} as an oracle. For each prompt, it compares a watermarked completion with an unwatermarked completion generated by the same model under identical settings. Each pair is judged twice with the positions swapped; a win is counted only when both orderings agree, while disagreements and equal-quality judgments are treated as ties. NPS is the difference between watermarked and unwatermarked wins divided by the total number of pairs, including ties. Positive values therefore favor watermarked completions, negative values favor unwatermarked completions, and values near zero indicate no clear preference. The exact query is provided in Appendix~\ref{app:prompt-tq}.

\subsection{Evaluation}
In unattacked settings, several schemes achieve near-saturated detection after 200 tokens. At 1\% FPR, \scheme{DEW} reaches TPRs of 98.6\% and 99.9\% for $\lambda=2.0$ and $\lambda=2.5$, respectively, placing it alongside the strongest semantic and surface-level schemes.

Across paraphrasing and translation, \scheme{DEW} shows strong robustness. At $\lambda=2.5$, its paraphrase TPRs are 54.3\% and 83.1\% at 1\% and 5\% FPR, respectively, placing it among the stronger semantic methods. \scheme{ATW} performs best under paraphrasing, with TPRs of 64.6\% and 89.0\%, and also outperforms \scheme{DEW} at $\lambda=2.0$ at approximately matched NPS. However, this advantage comes at substantially higher computational cost (Table~\ref{tab:comp-efficiency}). Translation robustness is the clearest strength of \scheme{DEW}: at $\lambda=2.5$, it achieves German TPRs of 52.1\% and 90.4\% and French TPRs of 29.9\% and 75.4\% at 1\% and 5\% FPR, respectively, the highest among the evaluated schemes.

Increasing $\lambda$ from 2.0 to 2.5 improves \scheme{DEW}'s TPR under all three transformations, while its NPS decreases from $-0.046$ to $-0.108$. Thus, the watermark strength provides a tunable robustness--quality trade-off. Overall, \scheme{DEW} combines competitive paraphrase robustness with the strongest post-translation detection in this evaluation, suggesting that dual-embedding watermarking is a promising substrate for semantic watermarks. A complementary analysis using human-authored English negatives is reported in Appendix~\ref{app:human-negatives}; \scheme{DEW}'s translation lead is stable, whereas the leading paraphrase method at 1\% FPR depends on the negative class. Appendix~\ref{app:supp-experiments} also contains supplementary experiments and analyses covering ablations, secrecy, additional attacks, and additional language models.

\subsection{Computational Efficiency}
\label{subsec:comp-efficiency}
\begingroup
\newcommand{\best}{\bfseries}
\newcommand{\underlinenum}[3]{{\phantom{\tablenum[table-format=#1,#2]{#3}}\llap{\underline{\num[#2]{#3}}}}}
\newcommand{\categorygen}[1]{\underlinenum{2.3}{}{#1}}
\newcommand{\categorydet}[1]{\underlinenum{1.4}{print-zero-integer=false}{#1}}
\newcommand{\genuncertainty}[1]{\,{\scriptsize\(\pm\)}\,{\scriptsize\tablenum[table-format=1.3]{#1}}}
\newcommand{\detuncertainty}[1]{\,{\scriptsize\(\pm\)}\,{\scriptsize\tablenum[table-format=1.4]{#1}}}
\newcommand{\gentime}[2]{\tablenum[table-format=2.3]{#1}\genuncertainty{#2}}
\newcommand{\dettime}[2]{\tablenum[table-format=1.4,print-zero-integer=false]{#1}\detuncertainty{#2}}
\begin{table}[t]
\centering
\caption{Mean per-text generation and detection runtimes in seconds, with half-ranges shown in smaller type. The lowest runtimes overall are bold, while the lowest runtimes in the other category (semantic/surface-level) are underlined.}
\label{tab:comp-efficiency}
\footnotesize
\setlength{\tabcolsep}{1.4pt}
\renewcommand{\arraystretch}{1.08}
\sisetup{detect-shape=true,detect-weight=true,mode=text,table-number-alignment=center}
\begin{tabular*}{\columnwidth}{@{\extracolsep{\fill}}lcc@{}}
\toprule
Watermark
& Generation (s)
& Detection (s)\\
\midrule
\multicolumn{3}{@{}l}{\textit{Semantic watermarks}} \\
\scheme{DEW}
& \gentime{4.822}{0.038} & \categorydet{0.0481}\detuncertainty{0.0003} \\
\scheme{k-SemStamp}
& \gentime{56.801}{1.309} & \dettime{0.0607}{0.0008} \\
\scheme{SIR}
& \gentime{7.151}{0.026} & \dettime{0.2825}{0.0007} \\
\scheme{X-SIR}
& \gentime{6.010}{0.030} & \dettime{0.1956}{0.0003} \\
\scheme{ATW}
& \gentime{10.156}{0.005} & \dettime{5.9967}{0.0182} \\
\scheme{TS}
& \categorygen{3.896}\genuncertainty{0.017} & \dettime{0.1150}{0.0002} \\
\addlinespace[.6ex]
\multicolumn{3}{@{}l}{\textit{Surface-level watermarks}} \\
\scheme{MorphMark}
& \gentime{3.925}{0.063} & \dettime{0.0350}{0.0003} \\
\scheme{KGW}
& {\best\tablenum[table-format=2.3]{3.800}}\genuncertainty{0.029} & \dettime{0.0379}{0.0002} \\
\scheme{SynthID}
& \gentime{4.251}{0.042} & {\best\tablenum[table-format=1.4,print-zero-integer=false]{0.0204}}\detuncertainty{0.0002} \\
\scheme{DiPmark}
& \gentime{3.906}{0.032} & \dettime{0.0598}{0.0002} \\
\scheme{UnbiasedWM}
& \gentime{3.954}{0.014} & \dettime{0.2340}{0.0010} \\
\midrule
\textit{(no watermark)}
& \textit{\tablenum[table-format=2.3]{3.790}} & -- \\
\bottomrule
\end{tabular*}
\end{table}
\endgroup

Table~\ref{tab:comp-efficiency} compares per-text generation and detection runtimes under the experimental setup in Section~\ref{sec:experiments} and Appendix~\ref{app:configs}. The experiments were conducted on an Intel i9-10980XE CPU and an NVIDIA RTX A5000 GPU, which was used for generation and for detectors that support GPU acceleration.

We use the publicly available MarkLLM implementations~\cite{pan_markllm_2024}, whose runtimes may differ from optimized deployments. While \scheme{ATW} is more robust to paraphrasing, its generation overhead is roughly six times \scheme{DEW}'s, and its detector is over two orders of magnitude slower in this implementation. Among the semantic watermarks, \scheme{DEW} has the fastest detection and is among the faster generators; only \scheme{TS} generates faster.

\section{Conclusion}
\label{sec:conclusion}
This paper presents \scheme{DEW}, a dual-embedding watermarking scheme designed for robustness to meaning-preserving text modifications. Our results show that dual-embedding watermarking can offer state-of-the-art robustness, particularly after translation. In our evaluation, \scheme{DEW} achieves the highest post-translation detection rates and remains competitive after paraphrasing. At lower watermark strength, it also maintains competitive text quality and incurs relatively low computational overhead compared with other semantic watermarks. These findings suggest that dual-embedding signals provide a promising substrate for future semantic watermarking schemes and motivate further study across models, languages, and deployment settings.

\section{Limitations}  %
\label{sec:limitations}
The reported operating points are calibrated on English unwatermarked completions. Their nominal false positive rates of 1\% and 5\% therefore do not imply the same false positive rates after paraphrasing or translation into German or French. Performance also depends on the watermark strength $\lambda$, the context and token embedding models, the host LLM, and the calibration domain. Our experiments cover a limited set of host models, prompt sets, languages, generation settings, and attacks; broader evaluation is needed to assess reliability across deployment settings.

The count-based watermark stealing experiment provides an initial assessment against one black-box spoofing attack that adapts to observed watermarked outputs, but is not tailored to \scheme{DEW}'s semantic projection. Broader adaptive analyses, including attacks targeting projection recovery, watermark removal, or secret recovery, remain important directions for future work.

The current \scheme{DEW} design leaves several choices open, including the projection dimensionality $n$, $\tanh$ scaling factor $\gamma$, embedding models, whitening transformations, and orthogonalization. Stronger or specialized embedding models may broaden language and host-model coverage. Finally, \scheme{DEW}'s applicability to instructed dialogue systems and low-entropy settings, including code generation, warrants further study, as do broader benchmarks and user studies assessing perceived quality, factual accuracy, creativity, and relevance.

\section*{Ethical Considerations}
\label{app:impact-statement}
This research advances watermarking methods for distinguishing LLM-generated text from human-authored content, with a focus on robustness to semantic transformations. More robust watermarks can support provenance and accountability when generated text is paraphrased or translated.

Deploying watermarks for LLM-generated text can support provenance and accountability, but it also risks false attribution of human-written text, overreliance in high-stakes moderation or legal settings, uneven reliability across languages and writing styles, and adversarial escalation through evasion, removal, or spoofing attacks.

We identify no substantive risks associated with the publication of our watermarking algorithm, and our contribution is purely methodological. The threat models we evaluate are standard in the watermarking literature and can be executed using publicly available tools. Consequently, disclosing our method does not introduce any new adversarial capabilities beyond those already well known in existing watermarking frameworks.

\section*{Acknowledgments}
This work was supported by the German Federal Ministry of Education and
Research (BMBF) under the programme \emph{Souverän. Digital. Vernetzt.} through
the joint project \emph{AIgenCY: Chances and Risks of Generative AI in
Cybersecurity} (project identification number 16KIS2013). GW was
additionally supported by the BMBF joint project \emph{6G-RIC: 6G Research and
Innovation Cluster} (project identification number 16KISK025).

\bibliography{references_cleaned_new_withYear}
\bibstyle{acl_natbib}

\newpage
\appendix

\onecolumn
\section{Geometry and Statistical Foundations of DEW}
\label{app:math}
\noindent\textbf{Reader map.}
Section~\ref{app:math:llm} frames next-token prediction as the composition of a \emph{context representation} map and a \emph{token scoring} (unembedding) map, and explains why \scheme{DEW} mirrors this structure via keyed signal processing on embeddings.
Section~\ref{app:math:dist} derives the null distributions of the core alignment score, emphasizing that the exact law is Beta-type while a Gaussian approximation emerges in high dimensions.
Section~\ref{app:math:test} states a concise one-sided hypothesis test for watermark detection and clarifies the approximation points.

\subsection{Inner-product geometry of next-token prediction}
\label{app:math:llm}

Let $x_{<t}$ denote the prefix and $w$ a candidate next token.
Decoder-only LLMs can be abstracted as two coupled maps: a \emph{context map} that builds a representation of the prefix, and a typically near-linear \emph{token scoring} (unembedding) map that produces next-token logits,
\begin{align*}
    h_t &= f_\theta(x_{<t}) \in \mathbb{R}^{d},\\
    \ell_t(w) &\approx \langle W_U[w],\, h_t\rangle + b_w,\\
    \mathbb{P}(w\mid x_{<t}) &= \mathrm{softmax}(\ell_t)(w).
\end{align*}
This factorization makes clear why inner products are a natural primitive for next-token selection.
Moreover, recent theory suggests that training pressure can make latent variables (``concept'' directions) linearly accessible in representation space~\cite{Jiang2024LinearRepresentations}. Thus, small perturbations expressed as controlled linear scores can interact smoothly with semantics and degrade gracefully under semantic shifts.

\scheme{DEW} mirrors this structure externally (without access to the model's internal residual stream) using embedding models and keyed linear maps.
For each position $i$, let $\mathbf{c}^{(i)}=(x_{i-w},\ldots,x_{i-1})$ be the context window and define
\begin{align*}
    \mathbf{e}_C^{(i)} &= M_C(\mathbf{c}^{(i)})\in\mathbb{R}^{d_C},\qquad
    \hat{\mathbf{e}}_C^{(i)}=\frac{\mathbf{e}_C^{(i)}}{\|\mathbf{e}_C^{(i)}\|},\\
    \mathbf{p}_C^{(i)} &= \frac{\hat{\mathbf{e}}_C^{(i)}\mathbf{R}_C}{\|\hat{\mathbf{e}}_C^{(i)}\mathbf{R}_C\|}\in\mathbb{R}^{n}.
\end{align*}
Similarly, let $\mathbf{e}_T^{(i)}\in\mathbb{R}^{d_T}$ be the whitened token embedding of $x_i$ and $\hat{\mathbf{e}}_T^{(i)}=\frac{\mathbf{e}_T^{(i)}}{\|\mathbf{e}_T^{(i)}\|}$; define
\begin{align*}
    \mathbf{p}_T^{(i)}=\frac{\hat{\mathbf{e}}_T^{(i)}\mathbf{R}_T}{\|\hat{\mathbf{e}}_T^{(i)}\mathbf{R}_T\|}\in\mathbb{R}^{n}.
\end{align*}
The core alignment score is the cosine similarity, expressed as the dot product
\begin{align*}
    Z^{(i)} := \mathbf{p}_T^{(i)}(\mathbf{p}_C^{(i)})^\top \in [-1,1],
\end{align*}
which is then mapped (monotonically, e.g., via $\tanh$) to a logit bias.
Hence, \scheme{DEW} can be viewed as a keyed signal-processing layer that injects a small, structured logit bias consistent with the inner-product geometry underlying next-token prediction.

\subsection{Underlying distributions of the alignment score}
\label{app:math:dist}

This section characterizes the baseline distribution of the dot product
\(
Z := \mathbf{p}_T\mathbf{p}_C^\top \in [-1,1]
\)
(and its scaled version $\sqrt{n}\,Z$), which underpins both watermark insertion and detection, and is used to calibrate false-positive control in Section~\ref{app:math:test}.

\paragraph{Setup and what is (approximately) known.}
Token embeddings are whitened in preprocessing, making $M_T(x)$ approximately isotropic for tokens $x\in\mathcal{V}$.
After a key-seeded random projection and normalization, it is reasonable to model $\mathbf{p}_T$ as approximately uniform on the unit sphere $\mathbf{S}^{n-1}$.
Context embeddings $\mathbf{e}_C=M_C(\mathbf{c})$ are not generally isotropic (they often lie in an anisotropic cone that reflects semantic constraints), so $\mathbf{p}_C$ need not be uniform.
Crucially, for this baseline distribution, it suffices that, conditional on the context projection $\mathbf{p}_C$, the token projection $\mathbf{p}_T$ is approximately uniform on $\mathbf{S}^{n-1}$.

\begin{assumption}[Conditionally spherical token projection (baseline)]\label{ass:spherical}
In the non-watermarked regime used for false-positive calibration, for each position $i$, conditional on the context projection $\mathbf{p}_C^{(i)}$, the normalized projected token vector satisfies $\mathbf{p}_T^{(i)}\approx \mathcal{U}(\mathbf{S}^{n-1})$.
\end{assumption}

\paragraph{Why Assumption~\ref{ass:spherical} is plausible under a fixed key.}
Although the secret key fixes $\mathbf{R}_T$ for all documents, randomness remains through the token sequence under $H_0$.
For a freshly sampled Gaussian projection and any fixed unit embedding vector, the projected vector is spherical before normalization.
In deployment, however, the key is fixed, so this key-averaged sphericality becomes an approximation over the empirical distribution of tokens and documents.
Whitening and normalization make this approximation more plausible by reducing dominant anisotropic directions in the token embeddings, but they do not make the fixed-key null exactly spherical.
Residual deviations from the spherical model can be handled by conservative calibration, e.g., using $L_{\mathrm{eff}}$ or empirical null estimation.

\paragraph{Exact law (Beta-type).}
\begin{lemma}[Dot product with a spherical vector]\label{lem:sphere_dot}
Let $Y\sim \mathcal{U}(\mathbf{S}^{n-1})$ and let $x\in\mathbf{S}^{n-1}$ be any fixed unit vector.
Then $x^\top Y$ has density
\begin{align*}
    f(z) = \frac{\Gamma\!\left(\frac{n}{2}\right)}{\sqrt{\pi}\,\Gamma\!\left(\frac{n-1}{2}\right)}(1-z^2)^{\frac{n-3}{2}},\qquad z\in[-1,1],
\end{align*}
Equivalently, $\frac{1+z}{2}\sim \mathrm{Beta}\!\left(\frac{n-1}{2},\frac{n-1}{2}\right)$.
In particular, $\mathbb{E}[x^\top Y]=0$ and $\mathrm{Var}(x^\top Y)=\frac{1}{n}$.
\end{lemma}
\smallskip
Applying the lemma conditionally with $Y=\mathbf{p}_T$ and $x=\mathbf{p}_C$ (treating $\mathbf{p}_C$ as fixed or slowly varying) yields an exact description of $Z$ in the baseline regime as long as $\mathbf{p}_T$ is (approximately) uniform on the sphere, regardless of whether $\mathbf{p}_C$ is anisotropic.

\begin{lemma}[High-dimensional Gaussian approximation]\label{lem:gauss_limit}
Under Assumption~\ref{ass:spherical}, let $Z := \mathbf{p}_T\mathbf{p}_C^\top\in[-1,1]$ with $\mathbf{p}_C\in\mathbf{S}^{n-1}$ treated as fixed (or conditioned upon).
Then, as $n\to\infty$,
\begin{align*}
    \sqrt{n}\,Z \;\xrightarrow{d}\; \mathcal{N}(0,1).
\end{align*}
\end{lemma}
\smallskip
Lemma~\ref{lem:gauss_limit} motivates scaling by $\sqrt{n}$ so that the per-token score has approximately unit variance in the baseline regime.

\subsection{One-sided hypothesis test for watermark detection}
\label{app:math:test}

\scheme{DEW}'s detector can be interpreted as a one-sided hypothesis test with false-positive control.
For brevity, we first present the linear (non-saturated) statistic; the $\tanh$ nonlinearity is discussed at the end.

Let $b_1,\dots,b_L$ be token-level bias scores for a document (Algorithm~\ref{alg:watermark_detection}):
\begin{align*}
    b_i \;=\; \lambda\sqrt{n}\,\bigl(\mathbf{p}_T^{(i)}(\mathbf{p}_C^{(i)})^\top\bigr)\in\mathbb{R},
    \qquad
    \bar{b} \;=\; \frac{1}{L}\sum_{i=1}^L b_i .
\end{align*}

\paragraph{Null $H_0$ (not watermarked).}
Under $H_0$, token selection is not influenced by the key.
Assumption~\ref{ass:spherical} formalizes the resulting spherical model for $\mathbf{p}_T^{(i)}$, and Lemma~\ref{lem:sphere_dot} yields the exact per-token dot-product law.
Using Section~\ref{app:math:dist} with $\mathbf{p}_T^{(i)}\approx \mathcal{U}(\mathbf{S}^{n-1})$, we have (conditionally on $\mathbf{p}_C^{(i)}$)
\begin{align*}
    \mathbb{E}[b_i]=0,\qquad \mathrm{Var}(b_i)=\lambda^2,
\end{align*}
and the exact single-token distribution is $\lambda\sqrt{n}\,Z$ where $Z$ is Beta-type on $[-1,1]$.
For document-level inference, we use a CLT approximation:
if $(b_i)$ are independent or weakly dependent with an effective sample size $L_{\mathrm{eff}}\le L$, then
\begin{align*}
    \bar{b} \;\approx\; \mathcal{N}\!\left(0,\frac{\lambda^2}{L_{\mathrm{eff}}}\right).
\end{align*}
(Practically, $L_{\mathrm{eff}}$ can be set to $L$ under an i.i.d.\ approximation, or conservatively reduced to account for correlations across nearby tokens.)

\paragraph{Alternative $H_1$ (watermarked with \scheme{DEW}).}
Under $H_1$, \scheme{DEW} biases token probabilities toward larger alignments $\mathbf{p}_T^{(i)}(\mathbf{p}_C^{(i)})^\top$, inducing a positive mean shift:
\begin{align*}
    \mathbb{E}[b_i]=\mu>0,
\end{align*}
and thus $\mathbb{E}[\bar{b}]=\mu>0$ while the variance remains comparable for small watermark strengths.

\paragraph{Test statistic and rejection rule.}
We test $H_0:\mu\le 0$ against $H_1:\mu>0$ using
\begin{align*}
    Z_L \;=\; \frac{\bar{b}}{\frac{\lambda}{\sqrt{L_{\mathrm{eff}}}}} \;=\; \frac{\sqrt{L_{\mathrm{eff}}}\,\bar{b}}{\lambda}.
\end{align*}
Under $H_0$, $Z_L\approx \mathcal{N}(0,1)$, and the one-sided p-value is $p=1-\Phi(Z_L)$.
At significance level $\alpha$, reject $H_0$ if $Z_L>z_\alpha$ (equivalently $p<\alpha$), where $z_\alpha$ is the $(1-\alpha)$-quantile of the standard normal.

\paragraph{Analytic classification threshold and empirical agreement.}
The rejection rule $Z_L>z_\alpha$ is equivalent to an analytic threshold on the document score,
\begin{align*}
    \bar{b} \;>\; \tau_\alpha
    \qquad\text{with}\qquad
    \tau_\alpha \;:=\; \frac{\lambda}{\sqrt{L_{\mathrm{eff}}}}\,z_\alpha,
\end{align*}
which yields a closed-form decision boundary for any target false-positive rate $\alpha$ under the Gaussian null approximation.

\begin{remark}[Bounded nonlinearity]\label{rem:bounded}
If the detector uses the saturated score $b_i=\lambda\tanh(\gamma\sqrt{n}\,\mathbf{p}_T^{(i)}(\mathbf{p}_C^{(i)})^\top)$, then $b_i$ is bounded and symmetric under $H_0$. Moreover, in the linear regime where $|\gamma\sqrt{n}\,Z|\ll 1$ and $\tanh(u)\approx u$, we have $b_i \approx \lambda\gamma\sqrt{n}\,Z$, hence $\mathrm{Var}(b_i)\approx (\lambda\gamma)^2$ under $H_0$.
The same test structure applies by replacing $\lambda^2$ with $\mathrm{Var}(b_i)$ under $H_0$, which can be estimated empirically (or approximated using the Gaussian limit for $\sqrt{n}\,Z$).
\end{remark}

\section{Orthogonal Construction of Random Matrices}
\label{app:orthogonal_construction}
In the following, we propose an optional block-wise row-orthonormal
construction of the random projection matrices \(\mathbf{R}_T\) and
\(\mathbf{R}_C\). Although such orthonormality is not mandatory for the
functionality of \scheme{DEW}, this construction preserves inner products,
norms, and therefore angles within each embedding space after projection.

We select the \emph{projection dimensionality} \(n\) as the least common
multiple of \(d_T\) and \(d_C\) to ensure the existence of integers
\(k_T\) and \(k_C\) such that
\begin{equation}
    \frac{n}{d_T} = k_T
    \quad \text{and} \quad
    \frac{n}{d_C} = k_C .
\end{equation}

This choice allows \(\mathbf{R}_T\) and \(\mathbf{R}_C\) to be constructed
by concatenating \(k_T\) and \(k_C\) square orthogonal blocks, respectively,
followed by an appropriate scaling. The resulting matrices satisfy
\[
    \mathbf{R}_T \mathbf{R}_T^\top = \mathbf{I}_{d_T}
    \quad \text{and} \quad
    \mathbf{R}_C \mathbf{R}_C^\top = \mathbf{I}_{d_C}.
\]
Thus, each projection is an isometric embedding into \(\mathbb{R}^n\).
In particular, if the input embeddings are normalized before projection,
then their projected representations are already normalized, so no
additional post-projection normalization is required for norm preservation.

We describe the construction for \(\mathbf{R}_C\); the construction for
\(\mathbf{R}_T\) is analogous.

\begin{itemize}
    \item Generate random matrices
    \(\widetilde{\mathbf{R}}_j \in \mathbb{R}^{d_C \times d_C}\)
    for \(j = 1, \ldots, k_C\).

    \item Orthonormalize each matrix, for example via QR decomposition
    or Gram--Schmidt, to obtain orthogonal matrices
    \(\mathbf{Q}_j \in \mathbb{R}^{d_C \times d_C}\) satisfying
    \[
        \mathbf{Q}_j \mathbf{Q}_j^\top
        =
        \mathbf{Q}_j^\top \mathbf{Q}_j
        =
        \mathbf{I}_{d_C}.
    \]

    \item Build
    \[
        \mathbf{R}_C
        \coloneq
        \frac{1}{\sqrt{k_C}}
        \bigl[
            \mathbf{Q}_1
            \mid
            \mathbf{Q}_2
            \mid
            \cdots
            \mid
            \mathbf{Q}_{k_C}
        \bigr]
        \in \mathbb{R}^{d_C \times n}.
    \]
    Equivalently, \(\mathbf{R}_C\) has orthonormal rows:
    \[
        \mathbf{R}_C \mathbf{R}_C^\top
        =
        \mathbf{I}_{d_C}.
    \]

    \item For any vector \(\mathbf{v} \in \mathbb{R}^{1 \times d_C}\),
    the projection is
    \[
        \mathbf{v}\mathbf{R}_C
        =
        \frac{1}{\sqrt{k_C}}
        \bigl[
            \mathbf{v}\mathbf{Q}_1
            \mid
            \mathbf{v}\mathbf{Q}_2
            \mid
            \cdots
            \mid
            \mathbf{v}\mathbf{Q}_{k_C}
        \bigr].
    \]
    Since each \(\mathbf{Q}_j\) is orthogonal,
    \(\|\mathbf{v}\mathbf{Q}_j\|_2=\|\mathbf{v}\|_2\).
    Therefore,
    \[
        \|\mathbf{v}\mathbf{R}_C\|_2^2
        =
        \frac{1}{k_C}
        \sum_{j=1}^{k_C}
        \|\mathbf{v}\mathbf{Q}_j\|_2^2
        =
        \|\mathbf{v}\|_2^2.
    \]

    \item More generally, for any
    \(\mathbf{u},\mathbf{v}\in\mathbb{R}^{1\times d_C}\),
    \[
        (\mathbf{u}\mathbf{R}_C)(\mathbf{v}\mathbf{R}_C)^\top
        =
        \frac{1}{k_C}
        \sum_{j=1}^{k_C}
        \mathbf{u}\mathbf{Q}_j\mathbf{Q}_j^\top\mathbf{v}^\top
        =
        \mathbf{u}\mathbf{v}^\top.
    \]
    Thus, right multiplication by \(\mathbf{R}_C\) is an isometric
    embedding from \(\mathbb{R}^{d_C}\) into \(\mathbb{R}^n\), preserving
    norms, inner products, and angles within the context-embedding space.
\end{itemize}

\begingroup
\newcommand{\uncertainty}[1]{{\scriptsize\,\(\pm\)\,#1}}
\newcommand{\best}{\bfseries}
\begin{table*}[b]
\centering
\caption{Semantic-mode ablation of \scheme{DEW} under English same-model calibration. Values are means across three runs, with half-ranges shown in smaller type. Best values are bold.}
\label{tab:ablation}
\footnotesize
\setlength{\tabcolsep}{1.4pt}
\renewcommand{\arraystretch}{1.10}
\sisetup{detect-shape=true,detect-weight=true,mode=text,table-number-alignment=center}
\begin{tabular*}{\linewidth}{@{\extracolsep{\fill}}l*{7}{S[table-format=1.3,print-zero-integer=false]@{}l}S[table-format=1.3]@{}lS[table-format=+1.3,print-zero-integer=false,retain-explicit-plus=true]@{}l@{}}
\toprule
& \multicolumn{2}{c}{Unattacked}
& \multicolumn{4}{c}{Paraphrased}
& \multicolumn{4}{c}{German translation}
& \multicolumn{4}{c}{French translation}
& \multicolumn{4}{c}{Text quality} \\
\cmidrule(lr){2-3}\cmidrule(lr){4-7}\cmidrule(lr){8-11}\cmidrule(lr){12-15}\cmidrule(l){16-19}
Semantic mode
& \multicolumn{2}{c}{TPR@1\%}
& \multicolumn{2}{c}{TPR@1\%} & \multicolumn{2}{c}{TPR@5\%}
& \multicolumn{2}{c}{TPR@1\%} & \multicolumn{2}{c}{TPR@5\%}
& \multicolumn{2}{c}{TPR@1\%} & \multicolumn{2}{c}{TPR@5\%}
& \multicolumn{2}{c}{PPL $\downarrow$} & \multicolumn{2}{c}{NPS $\uparrow$} \\
\midrule
\texttt{both} (default)
& 0.986 & \uncertainty{.008} & 0.352 & \uncertainty{.065} & 0.711 & \uncertainty{.019}
& \best 0.438 & \uncertainty{.066} & \best 0.857 & \uncertainty{.031} & \best 0.243 & \uncertainty{.069}
& \best 0.691 & \uncertainty{.039} & 8.938 & \uncertainty{.000} & -0.046 & \uncertainty{.029} \\
\texttt{context\_only}
& \best 0.999 & \uncertainty{.001} & \best 0.570 & \uncertainty{.022} & \best 0.797 & \uncertainty{.009}
& 0.043 & \uncertainty{.014} & 0.153 & \uncertainty{.022} & 0.025 & \uncertainty{.008}
& 0.083 & \uncertainty{.002} & 8.708 & \uncertainty{.063} & -0.073 & \uncertainty{.034} \\
\texttt{token\_only}
& \best 0.999 & \uncertainty{.001} & 0.123 & \uncertainty{.023} & 0.356 & \uncertainty{.003}
& 0.005 & \uncertainty{.001} & 0.047 & \uncertainty{.004} & 0.008 & \uncertainty{.000}
& 0.055 & \uncertainty{.005} & \best 8.667 & \uncertainty{.063} & \best +0.002 & \uncertainty{.029} \\
\texttt{neither}
& \best 0.999 & \uncertainty{.001} & 0.200 & \uncertainty{.023} & 0.394 & \uncertainty{.016}
& 0.009 & \uncertainty{.003} & 0.046 & \uncertainty{.002} & 0.012 & \uncertainty{.002}
& 0.049 & \uncertainty{.009} & 8.875 & \uncertainty{.094} & -0.023 & \uncertainty{.035} \\
\bottomrule
\end{tabular*}
\end{table*}
\endgroup

\twocolumn
\section{Supplementary Experiments}
\label{app:supp-experiments}

\subsection{Ablation Study}
\label{app:ablation-study}
\begin{minipage}{\linewidth}
To isolate the roles of token- and context-level semantics, we evaluate four variants of \scheme{DEW}:
\begin{itemize}
  \item \texttt{both}: unmodified \scheme{DEW} (baseline).
  \item \texttt{context\_only}: token semantics are removed by randomly permuting the whitened token projections at initialization.
  \item \texttt{token\_only}: context semantics are removed by replacing the context projection with a pseudo-random unit vector seeded by the context token IDs.
  \item \texttt{neither}: both ablations are applied simultaneously.
\end{itemize}
\end{minipage}

\medskip
\noindent Table~\ref{tab:ablation} reports results using the default \mpnet{} context encoder, $\lambda=2.0$, token-level score aggregation, and English same-model leave-one-seed-out calibration. The remaining hyperparameters follow Section~\ref{subsec:lms-hyperparams}. All four modes remain nearly saturated without attack, so their differences emerge primarily after transformation.

\paragraph{Paraphrasing.} The \texttt{context\_only} variant outperforms the full method at both operating points consistently across all three runs. Score diagnostics indicate that it begins with a stronger effective watermark at the same numerical $\lambda$ and remains more separable after paraphrasing, rather than retaining a larger fraction of its initial signal. We hypothesize that permuting the token projections changes the watermark-score distribution over likely candidate tokens, but do not test this mechanism directly.

\paragraph{Translation.} The full method retains substantially more signal after German and French translation than any ablated variant. Both token and context semantics are therefore needed for the full cross-lingual signal in this setting.

\paragraph{Context encoder.}
We compare the default \mpnet{}~\cite{reimers_sentence-bert_2019} context encoder with \efive{}~\cite{wang2024multilingual} at two watermark strengths. Table~\ref{tab:encoder-sensitivity} reports AUROC against same-model unwatermarked completions in the corresponding condition. We use AUROC because it directly measures how well each encoder separates watermarked from unwatermarked scores across the full range of detection thresholds.

\begingroup
\newcommand{\best}{\bfseries}
\newcommand{\uncertainty}[1]{{\scriptsize\,\(\pm\)\,#1}}
\newcommand{\aurocvalue}[2]{\tablenum[table-format=1.4]{#1}\uncertainty{#2}}
\newcommand{\npsvalue}[2]{\tablenum[table-format=-1.3]{#1}\uncertainty{#2}}
\begin{table*}[t]
\centering
\caption{Context-encoder sensitivity of \scheme{DEW} under token-level score aggregation. Values are means across three runs, with half-ranges shown in smaller type. Best values are bold.}
\label{tab:encoder-sensitivity}
\footnotesize
\setlength{\tabcolsep}{2.5pt}
\renewcommand{\arraystretch}{1.10}
\sisetup{detect-shape=true,detect-weight=true,mode=text,table-number-alignment=center}
\begin{tabular*}{\linewidth}{@{\extracolsep{\fill}}lcccccc@{}}
\toprule
& & \multicolumn{4}{c}{AUROC $\uparrow$} & \\
\cmidrule(lr){3-6}
Context encoder & $\lambda$ & Clean & Paraphrase & German & French & NPS $\uparrow$ \\
\midrule
MPNet & 2.0 & \aurocvalue{0.9994}{.0006} & \aurocvalue{0.9212}{.0071} & \aurocvalue{0.7079}{.0117} & \aurocvalue{0.7232}{.0093} & {\best\tablenum[table-format=-1.3]{-0.046}}\uncertainty{.029} \\
MPNet & 2.5 & \aurocvalue{0.9999}{.0001} & \aurocvalue{0.9537}{.0008} & \aurocvalue{0.7560}{.0101} & \aurocvalue{0.7650}{.0014} & \npsvalue{-0.108}{.029} \\
E5 & 2.0 & \aurocvalue{0.9998}{.0002} & \aurocvalue{0.9774}{.0014} & \aurocvalue{0.7936}{.0062} & \aurocvalue{0.7960}{.0070} & \npsvalue{-0.121}{.018} \\
E5 & 2.5 & {\best\tablenum[table-format=1.4]{1.0000}}\uncertainty{.0000} & {\best\tablenum[table-format=1.4]{0.9907}}\uncertainty{.0033} & {\best\tablenum[table-format=1.4]{0.8392}}\uncertainty{.0039} & {\best\tablenum[table-format=1.4]{0.8409}}\uncertainty{.0103} & \npsvalue{-0.217}{.039} \\
\bottomrule
\end{tabular*}
\end{table*}
\endgroup

At equal $\lambda$, E5 improves robustness, while its NPS is lower than MPNet's at the same strength. The closest observed quality points are MPNet at $\lambda=2.5$ and E5 at $\lambda=2.0$. Their robustness is broadly comparable, although E5 retains a modest AUROC advantage under each attack. Since these points are only approximately matched in quality, this does not establish encoder equivalence.

The same $\lambda$ need not yield the same effective watermark strength across context encoders and should therefore be selected separately under a fixed quality budget. We hypothesize that this difference results from encoder-dependent similarity geometry or isotropy, but the present experiments do not test this mechanism directly.

\subsection{Cross-Host Token Embedding Transfer}
\label{app:additional-llms}
We evaluate \scheme{DEW} on \gemmasevenb{}~\cite{DBLP:journals/corr/abs-2403-08295} and \falconsevenb{}~\cite{DBLP:journals/corr/abs-2311-16867} using the default \mpnet{}~\cite{reimers_sentence-bert_2019} context encoder, $\lambda=2.0$, and $\gamma=0.5$. The native configurations use each host model's input-embedding table and the corresponding whitening transform. In the donor configurations, each host token is decoded and re-tokenized with the \llamathreetwo{}~\cite{grattafiori_llama_2024} tokenizer, represented by the mean of the resulting input embeddings, and processed using the \llamathreetwo{} whitening transform. The host model, logits, and tokenizer used for generation remain unchanged.

All rows in Table~\ref{tab:cross-host-results} use token-level score aggregation and host-specific English same-model leave-one-seed-out calibration. Only the watermarked positives are paraphrased or translated. We report means and half-ranges across three runs.

\begingroup
\newcommand{\uncertainty}[1]{{\scriptsize\,\(\pm\)\,#1}}
\begin{table*}[t]
\centering
\caption{Cross-host comparison of native and fixed \llamathreetwo{} donor token tables under host-specific English same-model calibration. Values are means across three runs, with half-ranges shown in smaller type.}
\label{tab:cross-host-results}
\footnotesize
\setlength{\tabcolsep}{1.4pt}
\renewcommand{\arraystretch}{1.10}
\sisetup{detect-shape=true,detect-weight=true,mode=text,table-number-alignment=center}
\begin{tabular*}{\linewidth}{@{\extracolsep{\fill}}l*{7}{S[table-format=1.3,print-zero-integer=false]@{}l}S[table-format=2.3]@{}lS[table-format=+1.3,print-zero-integer=false,retain-explicit-plus=true]@{}l@{}}
\toprule
& \multicolumn{2}{c}{Unattacked}
& \multicolumn{4}{c}{Paraphrased}
& \multicolumn{4}{c}{German translation}
& \multicolumn{4}{c}{French translation}
& \multicolumn{4}{c}{Text quality} \\
\cmidrule(lr){2-3}\cmidrule(lr){4-7}\cmidrule(lr){8-11}\cmidrule(lr){12-15}\cmidrule(l){16-19}
Host
& \multicolumn{2}{c}{TPR@1\%}
& \multicolumn{2}{c}{TPR@1\%} & \multicolumn{2}{c}{TPR@5\%}
& \multicolumn{2}{c}{TPR@1\%} & \multicolumn{2}{c}{TPR@5\%}
& \multicolumn{2}{c}{TPR@1\%} & \multicolumn{2}{c}{TPR@5\%}
& \multicolumn{2}{c}{PPL $\downarrow$} & \multicolumn{2}{c}{NPS $\uparrow$} \\
\midrule
\multicolumn{19}{@{}l}{\textit{Native token embeddings}} \\
\gemmasevenb{}
& 0.981 & \uncertainty{.001} & 0.371 & \uncertainty{.054} & 0.673 & \uncertainty{.029}
& 0.039 & \uncertainty{.011} & 0.119 & \uncertainty{.012} & 0.179 & \uncertainty{.049}
& 0.487 & \uncertainty{.036} & 12.125 & \uncertainty{.188} & -0.051 & \uncertainty{.015} \\
\falconsevenb{}
& 0.995 & \uncertainty{.002} & 0.504 & \uncertainty{.049} & 0.707 & \uncertainty{.028}
& 0.056 & \uncertainty{.013} & 0.161 & \uncertainty{.012} & 0.088 & \uncertainty{.016}
& 0.277 & \uncertainty{.017} & 12.813 & \uncertainty{.094} & -0.021 & \uncertainty{.008} \\
\addlinespace[.6ex]
\multicolumn{19}{@{}l}{\textit{\llamathreetwo{} embeddings}} \\
\gemmasevenb{}
& 0.961 & \uncertainty{.005} & 0.341 & \uncertainty{.069} & 0.617 & \uncertainty{.048}
& 0.506 & \uncertainty{.060} & 0.792 & \uncertainty{.029} & 0.187 & \uncertainty{.049}
& 0.495 & \uncertainty{.048} & 12.313 & \uncertainty{.094} & +0.012 & \uncertainty{.014} \\
\falconsevenb{}
& 0.974 & \uncertainty{.012} & 0.336 & \uncertainty{.057} & 0.595 & \uncertainty{.049}
& 0.476 & \uncertainty{.078} & 0.808 & \uncertainty{.044} & 0.179 & \uncertainty{.026}
& 0.502 & \uncertainty{.043} & 12.625 & \uncertainty{.094} & -0.065 & \uncertainty{.014} \\
\bottomrule
\end{tabular*}
\end{table*}
\endgroup

The donor table substantially improves German translation retention on both hosts at both operating points. Its effect in French is host-dependent: the change is small on \gemmasevenb{} but more pronounced on \falconsevenb{}. Conversely, the native token tables retain more signal after paraphrasing, especially on \falconsevenb{}.

Text quality also changes in opposite directions across hosts. On \gemmasevenb{}, the donor configuration raises PPL but improves NPS, whereas on \falconsevenb{} it lowers PPL but reduces NPS. There is therefore no host-independent donor quality effect.

These results suggest that mapping host vocabularies into a fixed donor token space can improve translation retention across hosts. However, the benefit depends on the host and language and does not extend uniformly to paraphrasing or text quality. We therefore view donor-token construction as a promising but conditional transfer mechanism rather than a uniformly superior configuration.

\subsection{Sensitivity to Human-Authored English Negatives}
\label{app:human-negatives}
To examine sensitivity to the negative class, we repeat the primary evaluation using English human-authored texts instead of same-model unwatermarked completions for calibration. This is the only protocol change: Table~\ref{tab:human-negative-results} retains the same positives, attacks, configurations, quality measurements, token-level detection, and leave-one-seed-out procedure as Table~\ref{tab:results}.

The main translation result is stable under this alternative negative class, with \scheme{DEW} at $\lambda=2.5$ remaining the strongest method after German and French translation. The precise paraphrase ranking at 1\% FPR is sensitive to the negative class: \scheme{SIR} ranks first here, whereas \scheme{ATW} ranks first with same-model negatives; \scheme{ATW} remains strongest at 5\% FPR in both evaluations.

\begingroup
\newcommand{\uncertainty}[1]{{\scriptsize\,\(\pm\)\,#1}}
\newcommand{\best}{\bfseries}
\newcommand{\underlinenum}[3]{{\phantom{\tablenum[table-format=#1,#2]{#3}}\llap{\underline{\num[#2]{#3}}}}}
\newcommand{\categorytpr}[1]{\underlinenum{1.3}{print-zero-integer=false}{#1}}
\newcommand{\categoryppl}[1]{\underlinenum{2.2}{}{#1}}
\newcommand{\categorynps}[1]{\underlinenum{+1.3}{print-zero-integer=false,retain-explicit-plus=true}{#1}}
\begin{table*}[t]
\centering
\caption{True positive rates using English human-authored texts as the negative class for calibration and evaluation. Values are three-seed means, with half-ranges shown in smaller type. The best scores overall are bold, while the best scores in the other category are underlined.}
\label{tab:human-negative-results}
\footnotesize
\setlength{\tabcolsep}{1.4pt}
\renewcommand{\arraystretch}{1.10}
\sisetup{detect-shape=true,detect-weight=true,mode=text,table-number-alignment=center}
\begin{tabular*}{\textwidth}{@{\extracolsep{\fill}}l*{7}{S[table-format=1.3,print-zero-integer=false]@{}l}S[table-format=2.2]@{}lS[table-format=+1.3,print-zero-integer=false,retain-explicit-plus=true]@{}l@{}}
\toprule
& \multicolumn{2}{c}{Unattacked}
& \multicolumn{4}{c}{Paraphrased}
& \multicolumn{4}{c}{German translation}
& \multicolumn{4}{c}{French translation}
& \multicolumn{4}{c}{Text quality} \\
\cmidrule(lr){2-3}\cmidrule(lr){4-7}\cmidrule(lr){8-11}\cmidrule(lr){12-15}\cmidrule(l){16-19}
Watermark
& \multicolumn{2}{c}{TPR@1\%}
& \multicolumn{2}{c}{TPR@1\%} & \multicolumn{2}{c}{TPR@5\%}
& \multicolumn{2}{c}{TPR@1\%} & \multicolumn{2}{c}{TPR@5\%}
& \multicolumn{2}{c}{TPR@1\%} & \multicolumn{2}{c}{TPR@5\%}
& \multicolumn{2}{c}{PPL $\downarrow$} & \multicolumn{2}{c}{NPS $\uparrow$} \\
\midrule
\multicolumn{19}{@{}l}{\textit{Semantic watermarks}} \\
\scheme{DEW} ($\lambda=2.0$)
& 0.993 & \uncertainty{.007} & 0.430 & \uncertainty{.033} & 0.735 & \uncertainty{.019}
& 0.555 & \uncertainty{.013} & 0.882 & \uncertainty{.015} & 0.337 & \uncertainty{.019}
& 0.725 & \uncertainty{.020} & 8.94 & \uncertainty{0} & -0.046 & \uncertainty{.029} \\
\scheme{DEW} ($\lambda=2.5$)
& 0.999 & \uncertainty{.001} & 0.627 & \uncertainty{.024} & 0.843 & \uncertainty{.009}
& \best 0.625 & \uncertainty{.006} & \best 0.914 & \uncertainty{.002} & \best 0.393 & \uncertainty{.015}
& \best 0.787 & \uncertainty{.018} & 9.85 & \uncertainty{.06} & -0.108 & \uncertainty{.029} \\
\scheme{k-SemStamp}
& 0.835 & \uncertainty{.017} & 0.265 & \uncertainty{.006} & 0.519 & \uncertainty{.051}
& 0.003 & \uncertainty{.003} & 0.026 & \uncertainty{.005} & 0.012 & \uncertainty{.005}
& 0.037 & \uncertainty{.008} & {\categoryppl{8.00}} & \uncertainty{0} & \best +0.088 & \uncertainty{.037} \\
\scheme{SIR}
& 0.987 & \uncertainty{.004} & \best 0.733 & \uncertainty{.023} & 0.878 & \uncertainty{.010}
& 0.321 & \uncertainty{.046} & 0.591 & \uncertainty{.011} & 0.295 & \uncertainty{.032}
& 0.561 & \uncertainty{.015} & 9.79 & \uncertainty{.16} & -0.261 & \uncertainty{.029} \\
\scheme{X-SIR}
& 0.949 & \uncertainty{.002} & 0.701 & \uncertainty{.003} & 0.814 & \uncertainty{.003}
& 0.429 & \uncertainty{.011} & 0.615 & \uncertainty{.024} & 0.338 & \uncertainty{.016}
& 0.533 & \uncertainty{.010} & 9.54 & \uncertainty{.06} & -0.202 & \uncertainty{.040} \\
\scheme{ATW}
& \best 1.000 & \uncertainty{0} & 0.713 & \uncertainty{.012} & \best 0.907 & \uncertainty{.001}
& 0.020 & \uncertainty{.003} & 0.119 & \uncertainty{.004} & 0.003 & \uncertainty{.003}
& 0.026 & \uncertainty{.002} & 11.10 & \uncertainty{.13} & -0.048 & \uncertainty{.017} \\
\scheme{TS}
& \best 1.000 & \uncertainty{0} & 0.671 & \uncertainty{.022} & 0.799 & \uncertainty{.012}
& 0.167 & \uncertainty{.031} & 0.298 & \uncertainty{.022} & 0.092 & \uncertainty{.009}
& 0.172 & \uncertainty{.017} & 10.48 & \uncertainty{.06} & -0.125 & \uncertainty{.028} \\
\addlinespace[.6ex]
\multicolumn{19}{@{}l}{\textit{Surface-level watermarks}} \\
\scheme{MorphMark}
& 0.998 & \uncertainty{.002} & {\categorytpr{0.476}} & \uncertainty{.014} & {\categorytpr{0.699}} & \uncertainty{.023}
& 0.047 & \uncertainty{.016} & 0.132 & \uncertainty{.006} & {\categorytpr{0.075}} & \uncertainty{.017}
& 0.188 & \uncertainty{.014} & 8.63 & \uncertainty{0} & -0.062 & \uncertainty{.026} \\
\scheme{KGW}
& 0.999 & \uncertainty{.001} & 0.204 & \uncertainty{.035} & 0.387 & \uncertainty{.015}
& 0.026 & \uncertainty{.006} & 0.071 & \uncertainty{.015} & 0.020 & \uncertainty{.003}
& 0.067 & \uncertainty{.014} & 10.46 & \uncertainty{.16} & -0.113 & \uncertainty{.032} \\
\scheme{SynthID}
& 0.999 & \uncertainty{.001} & 0.351 & \uncertainty{.067} & 0.569 & \uncertainty{.016}
& 0.016 & \uncertainty{.008} & 0.051 & \uncertainty{.004} & 0.019 & \uncertainty{.005}
& 0.057 & \uncertainty{.008} & \best 6.68 & \uncertainty{.09} & {\categorynps{-0.002}} & \uncertainty{.060} \\
\scheme{DiPmark}
& 0.998 & \uncertainty{.002} & 0.139 & \uncertainty{.023} & 0.308 & \uncertainty{.013}
& 0.017 & \uncertainty{.011} & 0.069 & \uncertainty{.011} & 0.020 & \uncertainty{.008}
& 0.075 & \uncertainty{.012} & 9.15 & \uncertainty{.06} & -0.047 & \uncertainty{.018} \\
\scheme{UnbiasedWM}
& \best 1.000 & \uncertainty{0} & 0.183 & \uncertainty{.011} & 0.329 & \uncertainty{.013}
& {\categorytpr{0.129}} & \uncertainty{.030} & {\categorytpr{0.331}} & \uncertainty{.016} & 0.051 & \uncertainty{.009}
& {\categorytpr{0.199}} & \uncertainty{.009} & 9.19 & \uncertainty{.13} & -0.089 & \uncertainty{.039} \\
\bottomrule
\end{tabular*}
\end{table*}
\endgroup

\subsection{Robustness to Lexical Edits}
\label{app:superficial}
\begingroup
\begin{table*}[t]
\centering
\caption{True positive rates after word deletion and synonym substitution for the semantic watermark configurations evaluated in Table~\ref{tab:results}. English human-authored texts serve as the negative class for calibration and evaluation. The best scores are highlighted in bold.}
\label{tab:superficial-word-attacks}
\footnotesize
\resizebox{\textwidth}{!}{%
\begin{tabular}{lcccccccccccc}
\toprule
& \multicolumn{6}{c}{Word deletion} & \multicolumn{6}{c}{Synonym substitution} \\
\cmidrule(lr){2-7}\cmidrule(lr){8-13}
& \multicolumn{2}{c}{10\%} & \multicolumn{2}{c}{30\%} & \multicolumn{2}{c}{50\%}
& \multicolumn{2}{c}{10\%} & \multicolumn{2}{c}{30\%} & \multicolumn{2}{c}{50\%} \\
\cmidrule(lr){2-3}\cmidrule(lr){4-5}\cmidrule(lr){6-7}
\cmidrule(lr){8-9}\cmidrule(lr){10-11}\cmidrule(lr){12-13}
Watermark
& {TPR@1\%} & {TPR@5\%}
& {TPR@1\%} & {TPR@5\%}
& {TPR@1\%} & {TPR@5\%}
& {TPR@1\%} & {TPR@5\%}
& {TPR@1\%} & {TPR@5\%}
& {TPR@1\%} & {TPR@5\%} \\
\midrule
\scheme{DEW} ($\lambda=2.0$)
& 0.982 & \textbf{1.000} & 0.880 & 0.968 & 0.656 & 0.846
& 0.984 & 0.998 & 0.920 & 0.984 & 0.784 & 0.940 \\
\scheme{DEW} ($\lambda=2.5$)
& 0.998 & 0.998 & 0.964 & \textbf{0.994} & 0.852 & 0.946
& 0.998 & \textbf{1.000} & \textbf{0.990} & \textbf{0.998} & 0.920 & \textbf{0.984} \\
\scheme{k-SemStamp}
& 0.470 & 0.680 & 0.090 & 0.252 & 0.026 & 0.070
& 0.718 & 0.886 & 0.492 & 0.734 & 0.316 & 0.582 \\
\scheme{SIR}
& 0.966 & 0.984 & 0.940 & 0.978 & 0.904 & 0.944
& 0.966 & 0.990 & 0.932 & 0.984 & 0.878 & 0.952 \\
\scheme{X-SIR}
& 0.936 & 0.972 & 0.928 & 0.966 & \textbf{0.914} & \textbf{0.952}
& 0.920 & 0.962 & 0.864 & 0.944 & 0.824 & 0.914 \\
\scheme{ATW}
& 0.996 & \textbf{1.000} & 0.860 & 0.962 & 0.638 & 0.842
& \textbf{1.000} & \textbf{1.000} & 0.974 & \textbf{0.998} & 0.834 & 0.946 \\
\scheme{TS}
& \textbf{1.000} & \textbf{1.000} & \textbf{0.980} & 0.992 & 0.808 & 0.930
& \textbf{1.000} & \textbf{1.000} & \textbf{0.990} & 0.996 & \textbf{0.948} & 0.980 \\
\bottomrule
\end{tabular}}
\end{table*}
\endgroup

This auxiliary analysis uses the semantic watermark configurations from Table~\ref{tab:results} and reports single-seed point estimates without uncertainty intervals. Each cell evaluates 500 attacked watermarked texts using 500 English human-authored texts as the negative class for calibration and evaluation. Each scheme uses its own detector, and \scheme{DEW} uses token-level score aggregation.

Table~\ref{tab:superficial-word-attacks} evaluates robustness to lightweight lexical edits. Word deletion randomly removes whitespace-separated words with probability $r$, whereas context-aware synonym substitution first selects words with WordNet entries and then replaces masked positions with the top prediction of \bertlargeuncased{}. Thus, the latter should be interpreted as a contextual masked-token substitution rather than a strictly synonym-constrained transformation.

\scheme{DEW} remains highly robust across both lexical attacks. At $\lambda=2.5$, it achieves or ties the highest TPR in five of twelve columns and remains close to the leading semantic baseline in most others, including under 50\% edits. Although robustness decreases as the edit ratio increases, both \scheme{DEW} configurations generally achieve higher TPRs under contextual synonym substitution than under deletion.

\subsection{Robustness to DIPPER Paraphrasing}
\label{app:dipper-paraphrasing}
\begingroup
\begin{table*}[t]
\centering
\caption{True positive rates after DIPPER paraphrasing for the semantic watermark configurations evaluated in Table~\ref{tab:results}. English human-authored texts serve as the negative class for calibration and evaluation. The best value in each column is highlighted in bold.}
\label{tab:dipper_robustness}
\footnotesize
\setlength{\tabcolsep}{3.5pt}
\renewcommand{\arraystretch}{1.08}
\begin{tabular*}{\textwidth}{@{\extracolsep{\fill}}lcccccccc@{}}
\toprule
& \multicolumn{2}{c}{(\emph{ld}=40, \emph{od}=0)}
& \multicolumn{2}{c}{(\emph{ld}=40, \emph{od}=100)}
& \multicolumn{2}{c}{(\emph{ld}=60, \emph{od}=20)}
& \multicolumn{2}{c}{(\emph{ld}=60, \emph{od}=60)} \\
\cmidrule(lr){2-3}\cmidrule(lr){4-5}\cmidrule(lr){6-7}\cmidrule(lr){8-9}
Watermark
& {TPR@1\%} & {TPR@5\%}
& {TPR@1\%} & {TPR@5\%}
& {TPR@1\%} & {TPR@5\%}
& {TPR@1\%} & {TPR@5\%} \\
\midrule
\scheme{DEW} ($\lambda=2.0$)
& 0.734 & 0.916 & 0.416 & 0.688 & 0.462 & 0.726 & 0.314 & 0.572 \\
\scheme{DEW} ($\lambda=2.5$)
& 0.874 & 0.964 & 0.656 & 0.832 & 0.660 & 0.844 & 0.482 & 0.744 \\
\scheme{k-SemStamp}
& 0.406 & 0.672 & 0.220 & 0.462 & 0.266 & 0.472 & 0.172 & 0.368 \\
\scheme{SIR}
& 0.874 & 0.954 & \textbf{0.774} & \textbf{0.894} & \textbf{0.764} & 0.872 & \textbf{0.668} & \textbf{0.834} \\
\scheme{X-SIR}
& 0.798 & 0.870 & 0.730 & 0.826 & 0.706 & 0.812 & 0.666 & 0.790 \\
\scheme{ATW}
& \textbf{0.934} & \textbf{0.976} & 0.732 & 0.890 & 0.754 & \textbf{0.896} & 0.576 & 0.772 \\
\scheme{TS}
& 0.920 & 0.954 & 0.708 & 0.816 & 0.746 & 0.838 & 0.570 & 0.714 \\
\bottomrule
\end{tabular*}
\end{table*}
\endgroup

DIPPER~\cite{krishna_paraphrasing_2023} is an 11B-parameter paraphrase model trained to evade detectors for LLM-generated text, including watermarks. It uses surrounding context and controls lexical diversity and content reordering while aiming to preserve meaning.

This auxiliary analysis uses the semantic watermark configurations from Table~\ref{tab:results} and reports single-seed point estimates without uncertainty intervals. Each cell evaluates 500 DIPPER-paraphrased watermarked texts using 500 English human-authored texts as the negative class for calibration and evaluation, with token-level score aggregation for \scheme{DEW}. We denote the lexical- and order-diversity parameters by \emph{ld} and \emph{od}, respectively.

At $\lambda=2.5$, \scheme{DEW} is competitive at 5\% FPR, with TPRs within 0.012--0.090 of the strongest method across the four settings. At the stricter 1\% FPR, it ties \scheme{SIR} under $(\emph{ld}=40,\emph{od}=0)$ but generally trails the strongest schemes, particularly under the more aggressive paraphrasing settings.

\begin{table*}[b]
    \centering
    \caption{Evaluated configurations for the official MarkLLM implementations~\cite{pan_markllm_2024}.}
    \small
    \setlength{\tabcolsep}{5pt}
    \renewcommand{\arraystretch}{1.0}
    \label{tab:hyperparameters}
    \begin{tabular}{@{}p{0.16\textwidth}p{0.80\textwidth}@{}}
        \toprule
        \textbf{Watermark} & \textbf{Configuration} \\
        \midrule
        \scheme{$k$-SemStamp} & Authors' C4 SBERT encoder and C4 centroids; $k=8$, $\gamma=0.25$, margin $m=0.01$, $N_{\max}=100$, and 180--200 new tokens. \\
        \scheme{SIR} & Chunk length 10, $\delta=1.0$, compositional-BERT-large encoder, scale dimension 300, and the corresponding 128,256-token mapping. \\
        \scheme{X-SIR} & Chunk length 10, $\delta=1.0$, multilingual MPNet encoder, scale dimension 300, and the \llamathreetwo{} vocabulary mapping. \\
        \scheme{ATW} & Threshold 0.6, $\alpha=3.0$, top-$k=50$, top-$p=0.9$, repetition penalty 1.1, measure threshold 10, $\delta_0=0.2$, $\delta=0.35$, GPT-2-large measurement model, and MPNet embedding model. \\
        \scheme{TS} & $\gamma=0.5$, $\delta=2.0$, \texttt{simple\_1} seeding, prefix length 1, and the configured Llama checkpoint. \\
        \midrule
        \scheme{MorphMark} & Exponential scaling, $\gamma=0.5$, $k_{\mathrm{exp}}=1.3$, $p_0=0.15$, EWD disabled, prefix length 1, time-based seeding, and a left context window. \\
        \scheme{SynthID} & Non-distortionary mode, $3$-gram contexts, 30 fixed keys, sampling-table size 65,536, two leaves, context-history size 1,024, and mean-score detection. \\
        \scheme{KGW} & Prefix length 3, $\gamma=0.5$, $\delta=2.0$, time-based seeding, and a left context window. \\
        \scheme{DiPmark} & $\alpha=0.45$, $\gamma=0.5$, prefix length 3, and repeated contexts retained during generation and detection. \\
        \scheme{UnbiasedWM} & Gamma reweighting, grid size 10, prefix length 3, and repeated contexts ignored during generation and detection. \\
        \bottomrule
    \end{tabular}
\end{table*}

\subsection{Secrecy Evaluation via Watermark Stealing}
\label{app:secrecy}
\label{app:stealing}

We evaluate secrecy against the watermark stealing attack of \citet{jovanovic_watermark_2024}, as implemented in MarkLLM~\cite{pan_markllm_2024}. The attack operates in a black-box spoofing setting: the adversary observes text generated by a victim watermarked language model and estimates token-level continuation patterns that distinguish watermarked from unwatermarked generations. These estimates are then used to reweight the logits of an attacker-controlled language model, producing new texts that are intended to be accepted by the victim's detector.

We evaluate the \scheme{DEW} $\lambda=2.0$ configuration from Table~\ref{tab:results}. We generate a stolen corpus of 2000 watermarked completions with 200 tokens each and use the same base language model for the victim and attacker. The stealing model conditions its estimated token biases on the three preceding tokens. This auxiliary analysis reports single-seed point estimates without uncertainty intervals. We measure the TPR on 200 stolen-generated texts using 200 held-out English human-authored texts as the negative class for calibration and evaluation, with token-level \scheme{DEW} detection.

At 1\% and 5\% FPR, the stolen-generated texts achieve TPRs of 2.5\% and 5.0\%, respectively. The corresponding genuine-watermarked controls achieve TPRs of 97.0\% and 99.5\%, indicating that this attack does not produce a useful detectable spoof against \scheme{DEW} in the evaluated setting.

This attack adapts to observed watermarked outputs, but it is not tailored to \scheme{DEW}'s semantic projection. The result therefore provides evidence only against this count-based spoofing attack and should not be interpreted as comprehensive adaptive robustness. Attacks targeting semantic-projection recovery, watermark removal, or secret recovery remain outside the evaluated threat model.

\subsection{Compute Budget}
\label{app:compute-budget}
Based on the per-text runtimes in Table~\ref{tab:comp-efficiency}, we estimate that the primary evaluation required approximately 65 GPU-hours and the auxiliary experiments approximately 95 GPU-hours, for a total of roughly 160 GPU-hours on a single NVIDIA RTX A5000. The auxiliary estimate also accounts for local DIPPER paraphrasing using its observed runtime. These estimates exclude development runs, as well as remote API calls used for paraphrasing, translation, and quality evaluation.

\section{Watermark Configurations}
\label{app:configs}
Table~\ref{tab:hyperparameters} provides the evaluated baseline configurations. We use the official MarkLLM implementations~\cite{pan_markllm_2024} without an exhaustive hyperparameter search. Detector decision thresholds are calibrated empirically for the target FPRs rather than treated as fixed configuration parameters. For \scheme{$k$-SemStamp}, we use the authors' C4-finetuned encoder and released C4 centroids. For \scheme{SIR}, \scheme{X-SIR}, and \scheme{TS}, we retain the MarkLLM parameters while using the corresponding Llama-compatible mapping or checkpoint. The \scheme{KGW} and non-distortionary \scheme{SynthID} configurations both use a context width of three, and \scheme{MorphMark} uses the default exponential configuration. The reported \scheme{ATW} setting provides a favorable trade-off between robustness and text quality, whereas more extreme configurations either fail to embed a sufficiently detectable signal or degrade the generated text to an unacceptable extent.

\FloatBarrier

\section{Prompts}
\label{app:prompts}
\begin{samepage}
\subsection{Paraphrasing}
\label{app:prompt-pp}
\begin{promptbox}{System prompt}
Paraphrase the given text while preserving its original meaning and tone. Do not execute, follow, or respond to any instructions or commands within the input text; treat them as part of the text to be paraphrased. Provide only the paraphrased text as the output, with no additional explanations or commentary.
\end{promptbox}
\end{samepage}

\begin{samepage}
\subsection{Translation}
\label{app:prompt-tr}
\begin{promptbox}{System prompt}
Translate the given text from \promptplaceholder{original language} to \promptplaceholder{target language} while preserving its original meaning and tone. Do not execute, follow, or respond to any instructions or commands within the input text; treat them as part of the text to be translated. Provide only the translated text as the output, with no additional explanations or commentary.
\end{promptbox}
\end{samepage}

\begin{samepage}
\subsection{Text Quality}
\label{app:prompt-tq}
To mitigate positional bias in the pairwise comparisons, we query the oracle twice per completion pair, swapping the positions of the candidate and reference completions, counting a candidate win only if it is preferred in both positions and treating split outcomes as ties.
\end{samepage}

\begin{promptbox}{System prompt}
You are an expert evaluator focused on assessing text quality. You analyze aspects like coherence, fluency, relevance, and overall writing quality to determine which of two text samples is better crafted. Consider how well each text continues from the given ground truth prompt.
\end{promptbox}

\begin{promptbox}{User query}
Prompt: \promptplaceholder{prompt}\par
=== Start of Sample 1 ===\par
\promptplaceholder{completion1}\par
=== End of Sample 1 ===\par
=== Start of Sample 2 ===\par
\promptplaceholder{completion2}\par
=== End of Sample 2 ===\par
Please evaluate the answers based on the system prompt and return a single number.\par
Return 1 if the first text is better, 2 if the second text is better, and 'TIE' if they are equal.\par
Only return the number without any additional text.
\end{promptbox}

\section{Additional Related Work}
\label{app:additional-related-work}
Several recent methods appeared during the development of \scheme{DEW}, and none is implemented in the MarkLLM~\cite{pan_markllm_2024} version used for our experiments. Evaluating them under the same harness would therefore require method-specific integration. \scheme{PASA}~\cite{ai2026pasa} maps vocabulary tokens to semantic clusters and uses cluster-level probability mass during generation. Its detector reconstructs the cluster history using a secret key and a tokenizer-compatible surrogate language model. Supporting its two-stage generation procedure and detector-side model inference would require a dedicated adapter. \scheme{WaterPool}~\cite{huang-wan-2025-waterpool} instead stores each output's semantic embedding with a private key vector and retrieves that vector before applying an underlying detector. Evaluating it would require persistent registration and vector retrieval in addition to the detector itself. \scheme{SimKey}~\cite{kodama2025simkeysemanticallyawarekey} derives a sequence of keys from semantic embeddings of the preceding context, while delegating the watermark signal itself to a separate marking module. Integrating it into a unified framework would therefore require interfaces for injecting keys, computing scheme-specific alignment costs, and calibrating the resulting scores.

\FloatBarrier
\makeatletter
\setlength{\@dblfptop}{0pt}
\makeatother
\begin{table*}[!t]
\section{Artifact Licenses}
\label{app:artifact-licenses}
\centering
\caption{Licenses for major artifacts used in this work.}
\begin{tabular}{ll}
\hline
\textbf{Artifact} & \textbf{License} \\
\hline
\texttt{allenai/c4} & ODC-BY; subject to Common Crawl terms \\
\texttt{all-mpnet-base-v2} & Apache License 2.0 \\
\texttt{bert-large-uncased} & Apache License 2.0 \\
\texttt{compositional-bert-large-uncased} & Apache License 2.0 \\
\texttt{DIPPER} & Apache License 2.0 \\
\texttt{Falcon-7B} & Apache License 2.0 \\
\texttt{Gemma-7B} & Gemma Terms of Use \\
\texttt{gpt2-large} & Modified MIT License \\
\texttt{GPT-4o-mini} & Proprietary OpenAI API service terms \\
\texttt{Llama-3.1-8B} & Llama 3.1 Community License \\
\texttt{Llama-3.2-3B} & Llama 3.2 Community License \\
\texttt{MarkLLM} & Apache License 2.0 \\
\texttt{multilingual-e5-base} & MIT License \\
\texttt{paraphrase-multilingual-mpnet-base-v2} & Apache License 2.0 \\
\texttt{PyTorch} & BSD-style license \\
\texttt{Transformers} & Apache License 2.0 \\
\hline
\end{tabular}

\label{tab:artifact-licenses}
\end{table*}

\end{document}